\documentclass{article}
\PassOptionsToPackage{numbers,sort&compress}{natbib}

\usepackage[final]{neurips_2025}

\usepackage[utf8]{inputenc} % allow utf-8 input
\usepackage[T1]{fontenc}    % use 8-bit T1 fonts
\usepackage[colorlinks=true,linkcolor=blue,citecolor=blue,urlcolor=blue]{hyperref}
\usepackage{url}            % simple URL typesetting
\usepackage{booktabs}       % professional-quality tables
\usepackage{multirow}
\usepackage{graphicx}
\usepackage{subcaption}
\usepackage{amsfonts}       % blackboard math symbols
\usepackage{amsmath}
\usepackage{nicefrac}       % compact symbols for 1/2, etc.
\usepackage{microtype}      % microtypography
\usepackage{xcolor}         % colors

\usepackage{booktabs}
\usepackage{color, colortbl}
\definecolor{greyC}{RGB}{180,180,180}
\definecolor{greyL}{RGB}{235,235,235}
\definecolor{shadecolor}{rgb}{0.92,0.92,0.92}
\definecolor{color1}{RGB}{255,190,122}
\definecolor{color2}{RGB}{142,207,201}
\definecolor{color3}{RGB}{190,184,220}
\definecolor{color4}{RGB}{130,176,210}
\usepackage{multicol}
\usepackage{makecell}
\usepackage{algorithmic,algorithm}
\usepackage{soul}
\usepackage{threeparttable}
\usepackage{dsfont}
\usepackage{enumerate}
\usepackage{amsthm,amssymb}
\usepackage{wrapfig}
\usepackage{framed}
\usepackage{ulem}
\usepackage{rotating}
\usepackage{tablefootnote}
\usepackage{bbm}
\title{Preference-Driven Multi-Objective Combinatorial Optimization with Conditional Computation}

\author{
  Mingfeng Fan \\
  National University of Singapore\\
  \texttt{ming.fan@nus.edu.sg} \\
  \And
  Jianan Zhou\thanks{Corresponding author.}\\
  Nanyang Technological University \\
  \texttt{jianan004@e.ntu.edu.sg} \\
  \And
  Yifeng Zhang \\
  National University of Singapore \\
  \texttt{yifeng@u.nus.edu} \\
  \And
  Yaoxin Wu$^*$ \\
  Eindhoven University of Technology \\
  \texttt{y.wu2@tue.nl} \\
  \And
  Jinbiao Chen \\
  Sun Yat-sen University\\
  \texttt{chenjb69@mail2.sysu.edu.cn} \\
  \And
  Guillaume Adrien Sartoretti \\
  National University of Singapore\\
  \texttt{guillaume.sartoretti@nus.edu.sg} \\
}

\begin{document}

\maketitle

\begin{abstract}
  % Recent divide-and-conquer deep reinforcement learning methods have achieved remarkable success in solving multi-objective combinatorial optimization problems (MOCOPs) by decomposing them into multiple subproblems, each associated with a specific weight vector. However, existing approaches typically treat all subproblems equally by utilizing a single model, and fail to explore the vast solution space effectively, leading to inefficient learning and suboptimal performance. To address these limitations, we propose POCCO, a novel plug-and-play framework that allows subproblems to adaptively select model structures and optimize based on preference signals rather than explicit reward values. Specifically, we design a conditional computation block, enabling dynamic routing of subproblems to tailored neural architectures. Additionally, we develop a preference-driven optimization algorithm that reparameterizes the reward function using pairwise preference signals between winning and losing solutions, based on the Bradley–Terry model. We demonstrate the effectiveness and versatility of POCCO by integrating it with two state-of-the-art neural approaches to solve MOCOPs. Experimental results across four classic MOCOP benchmarks show that POCCO significantly outperforms baselines and exhibits superior generalization capabilities.
  Recent deep reinforcement learning methods have achieved remarkable success in solving multi-objective combinatorial optimization problems (MOCOPs) by decomposing them into multiple subproblems, each associated with a specific weight vector. However, these methods typically treat all subproblems equally and solve them using a single model, hindering the effective exploration of the solution space and thus leading to suboptimal performance. To overcome the limitation, we propose POCCO, a novel plug-and-play framework that enables adaptive selection of model structures for subproblems, which are subsequently optimized based on preference signals rather than explicit reward values. Specifically, we design a conditional computation block that routes subproblems to specialized neural architectures. Moreover, we propose a preference-driven optimization algorithm that learns pairwise preferences between winning and losing solutions. We evaluate the efficacy and versatility of POCCO by applying it to two state-of-the-art neural methods for MOCOPs. Experimental results across four classic MOCOP benchmarks demonstrate its significant superiority and strong generalization.  
\end{abstract}

\section{Introduction}
\label{sec:intro}
Multi-objective combinatorial optimization problems (MOCOPs) involve optimizing multiple conflicting objectives within a discrete decision space. They have attracted considerable attention from the computer science and operations research communities due to their widespread applications in manufacturing~\cite{ahmadi2016multi}, logistics~\cite{jozefowiez2008multi}, and scheduling~\cite{ghoseiri2004multi}. In such scenarios, decision makers must simultaneously consider and balance multiple criteria, such as cost, makespan, and environmental impact. Unlike single-objective combinatorial optimization problems (SOCOPs), which seek a single optimal solution, MOCOPs aim to identify Pareto optimal solutions that reflect trade-offs among conflicting objectives, making them inherently more challenging. Given their NP-hard nature, exact methods typically struggle to solve MOCOPs within reasonable time frames, as computational complexity may increase exponentially with problem scale~\cite{ehrgott2005multicriteria,florios2014generation}. Consequently, heuristic approaches have emerged as the main avenue for tackling MOCOPs. However, these heuristics often involve extensive iterative local searches for each new instance, resulting in high computational costs. Furthermore, conventional heuristics often rely on extensive domain-specific expertise and meticulous, problem-specific tuning, thus limiting their adaptability to broader classes of MOCOPs.

Recently, neural methods have achieved great success in solving SOCOPs~\cite{bello2016neural,nazari2018reinforcement,chen2019learning,hottung2021efficient,wu2021learning,kim2021learning,li2023learning,zhou2023towards,zhou2024mvmoe} by learning effective patterns of decision policies in a data-driven way. Motivated by these advances, researchers have extended neural approaches to MOCOPs, leveraging their advantages in bypassing labor-intensive heuristic design, accelerating problem solving through GPU parallelization, and flexibly adapting to diverse MOCOP variants.
Typically, neural methods address MOCOPs by decomposing them into a set of scalarized subproblems, each a SOCOP defined by a specific weight vector, and solving them using deep reinforcement learning (DRL) to approximate the Pareto front.
Early approaches train or fine-tune a separate model for each subproblem using transfer learning or meta-learning techniques~\cite{likaiwen2020deep, zhang2022meta}. However, these approaches require extensive computational resources and struggle to generalize to subproblems with unseen weight vectors.
As an alternative, PMOCO~\cite{lin2022pareto} employs a weight-conditioned hypernetwork to modulate model parameters, allowing a single model to address all subproblems. Nevertheless, it remains limited in effectively handling subproblems with diverse weight vectors.
More recently, methods such as CNH~\cite{fan2024conditional} and WE-CA~\cite{chen2025rethinking} have tackled this challenge by encoding weight vectors directly into the problem representations, resulting in a unified model that generalizes across various problem sizes. These methods are currently considered state-of-the-art (SOTA) in solving MOCOPs. 

% \begin{wrapfigure}{r}{0.4\textwidth}
%     \centering
%     \vspace{1mm}
%     \includegraphics[width=\linewidth]{scalarized_reward_bar.pdf}
%     \caption{\colorb{Scalarized rewards (i.e., negative solution costs) of MOTSP100 acorss different weight vectors.}}
%     \label{fig:example}
% \end{wrapfigure}

Current SOTA methods typically rely on a single neural network with limited capacity to handle all subproblems, which overcomplicates the learning task and results in suboptimal performance. A straightforward solution to ease training and promote effective representation learning across subproblems is to increase the model capacity. However, determining \emph{how much} additional capacity to allocate and \emph{where} to introduce it within the architecture remains an open challenge.
On the other hand, neural methods often adopt REINFORCE~\cite{williams1992simple} as the training algorithm, relying solely on scalarized objective values as reward signals to guide policy updates. Given its on-policy nature, REINFORCE suffers from high gradient variance and lacks structured mechanisms for effective exploration~\cite{ladosz2022exploration}. These issues are exacerbated in MOCOP settings, where the vast combinatorial action space makes efficient exploration particularly difficult, ultimately hindering policy performance.

To address these issues, we propose POCCO (\underline{P}reference‑driven multi‑objective combinatorial \underline{O}ptimization with \underline{C}onditional \underline{CO}mputation), a plug‑and‑play framework that augments neural MOCOP methods with two complementary mechanisms. 
First, POCCO introduces a \emph{conditional computation block} into the decoder, where a sparse gating network dynamically routes each subproblem through either a selected subset of feed-forward (FF) experts or a parameter-free identity (ID) expert. This design enables subproblems to adaptively select computation routes (i.e., model structures) based on their context, efficiently scaling model capacity and facilitating more effective representation learning.
Second, POCCO replaces raw scalarized rewards with \emph{pairwise preference learning}. For each subproblem, the policy samples two trajectories, identifies the better one as the winner, and maximizes a Bradley–Terry (BT) likelihood based on the difference in their average log-likelihoods.
% \colorb{This design inherently encourages broader exploration, as the gradient signal is strongest when the model is uncertain (i.e., the two trajectories are similarly preferred), which drives the policy to focus on ambiguous regions and improves learning efficiency \cite{rafailov2023direct}.}
Such comparative feedback guides the search toward policies that generate increasingly preferred solutions, enabling exploration of the most promising regions of the search space and more efficient convergence to higher-quality solutions.
% \colorr{Such comparative feedback empirically improves performance over REINFORCE by inherently guiding the search toward policies that generate increasingly preferred solutions.}
% First, a conditional computation block is inserted in the decoder: a sparse gating network routes every subproblem through either a selected subset of feed‑forward(FF) experts or a parameter‑free identity path, so each subproblem could choose its own computation paths. 
% Second, POCCO replaces raw scalarized rewards with pairwise preference learning: for each scalarized instance, the policy samples two trajectories, labels the better one as the winner, and maximizes a Bradley–Terry likelihood based on the difference in their average log‑likelihoods.  This scale‑invariant objective neutralizes reward‑magnitude disparities and encourages broader exploration. 
Our contributions are summarized as follows:
\begin{itemize}
    \item Conceptually, we address two fundamental limitations of existing approaches for solving MOCOPs: limited exploration within the vast solution space and the reliance on a single, capacity-limited model, which can lead to inefficient learning and suboptimal performance.
    \item Technically, we propose a conditional computation block that dynamically routes subproblems to tailored neural architectures. Additionally, we develop a preference-driven algorithm leveraging implicit rewards derived from pairwise preference signals between winning and losing solutions, modeled using the BT framework.
    \item Experimentally, we demonstrate the effectiveness and versatility of POCCO on classical MOCOP benchmarks using two SOTA neural methods. Extensive results show that POCCO not only outperforms all baseline methods but also exhibits superior generalization across diverse problem sizes.
\end{itemize}
% 1) Conceptually, we address two fundamental limitations of existing approaches for solving MOCOPs: limited exploration within the vast solution space and the reliance on a single, capacity-limited model, which can lead to inefficient learning and suboptimal performance.
% 2) Technically, we propose a conditional computation block that dynamically routes subproblems to tailored neural architectures. Additionally, we develop a preference-driven algorithm leveraging implicit rewards derived from pairwise preference signals between winning and losing solutions, modeled using the BT framework.
% 3) Experimentally, we demonstrate the effectiveness and versatility of POCCO on classical MOCOP benchmarks using two SOTA neural methods. Extensive results show that POCCO not only outperforms all baseline methods but also exhibits superior generalization across diverse problem sizes.

\section{Related Works}
\label{sec:review}
\textbf{Traditional methods for MOCOP.}
MOCOPs are significantly harder to solve than their single‑objective counterparts. Classic cases such as the Traveling Salesman Problem (TSP) and Capacitated Vehicle Routing Problem (CVRP) are already NP‑hard, and adding multiple objectives only magnifies the difficulty~\cite{florios2014generation}. Exact algorithms quickly become impractical for large instances or for problems that possess a vast Pareto set~\cite{halffmann2022exact,wu2022graph}. Consequently, research has shifted toward heuristic methods that deliver high‑quality Pareto fronts within reasonable time limits~\cite{zajac2021objectives,jozefowiez2008multi}. Among these, multi‑objective evolutionary algorithms (MOEAs) are widely used. They fall into two main categories: dominance‑based MOEAs~\cite{deb2002fast,deb2013evolutionary,seada2015unified} and decomposition‑based MOEAs~\cite{zhang2007moea,fang2020mining,ke2014simple}. Despite their popularity, MOEAs typically demand extensive manual design: practitioners must select and tune crossover, mutation, and selection operators along with many hyperparameters~\cite{yi2020behavior,xie2022dynamic,tian2021evolutionary}, and this labor‑intensive hand‑engineering often limits overall performance.

\textbf{Neural methods for MOCOP.}
Inspired by the success of DRL in SOCOPs \cite{berto2023rl4co}, recent research extends neural methods to MOCOPs by solving a series of scalarized SOCOPs. These neural solvers generally follow two paradigms: \emph{one-to-one} and \emph{one-to-many}.
The one-to-one paradigm trains or fine-tunes a separate neural model for each scalarized subproblem. Some approaches apply DRL algorithms individually and transfer parameters across models to accelerate convergence~\cite{wu2020modrl,likaiwen2020deep}. Others adopt architectures such as Pointer Networks~\cite{vinyals2015pointer} and Attention Models~\cite{kool2018attention,kwon2020pomo}, and optimize them using evolutionary strategies~\cite{shao2021multi,zhang2021modrl}. To promote generalization across subproblems, meta-learning techniques are introduced~\cite{zhang2022meta}. However, this paradigm suffers from high training overhead and the burden of maintaining multiple models.
In contrast, the one-to-many paradigm employs a single neural network to handle all subproblems. This includes hypernetwork-based DRL frameworks that condition on weight vectors~\cite{lin2022pareto,li2023deep,wu2023collaborative}, and conditional neural heuristics that incorporate instance features, weight information, and problem size~\cite{fan2024conditional, chen2025rethinking}.
% Some methods enhance diversity by incorporating hypervolume indicators into the reward~\cite{chen2024neural}, while others improve representation learning through conditional attention mechanisms~\cite{chen2025rethinking}. 
% or by fusing graph and image representations of problem instances~\cite{chen2025neural}.

\textbf{Preference optimization.}
Preference optimization methods, such as Direct Preference Optimization (DPO) \cite{rafailov2023direct} and Identity Preference Optimization (IPO) \cite{azar2024general}, have gained significant traction, particularly in large language model (LLM) training, due to their ability to directly optimize human preferences without relying on explicit reward modeling as required in reinforcement learning from human feedback (RLHF). These methods are typically designed for pairwise preference data, where human annotators identify a preferred output over a less preferred one in response to a given prompt.
Another line of research explores simpler preference optimization objectives that do not depend on a reference model~\cite{hong2024orpo, xu2023some}. Among them, SimPO~\cite{meng2024simpo} proposes using the average log-probability of a generated sequence as an implicit reward. This approach aligns more closely with the model’s generation process and improves computational and memory efficiency by eliminating the need for a separate reference model. Some studies~\cite{liao2025bopo} have explored applying preference optimization to SOCOPs, but its application to MOCOPs has been rarely investigated.

\section{Preliminary}
\label{sec:preliminary}

\subsection{MOCOP}
% \textbf{Decomposition-based Method.} 
% Typically, a VRP instance is defined on a graph $\mathcal{G}$ with the node set $\mathcal{V}=\left\{1, 2, \cdots, n\right\}$, where each node $i\!\in\! \mathcal{V}$ is featured by $o_{i}$. The solution to a VRP instance is a tour $\pi=\left(\pi_{1}, \pi_{2}, \cdots, \pi_{T}\right)$, i.e., a node sequence of length $T$, with $\pi_{j}\!\in \! \mathcal{V}$.
An MOCOP instance specified by data $\mathcal{G}$ can be formally defined as $\min_{\pi\in \mathcal{X}} F(\pi) = (f_{1}(\pi), f_{2}(\pi),\dots, f_{\kappa}(\pi))$, where $F$ is the objective vector with $\kappa$ objective functions, $\pi$ is a feasible solution, and $\mathcal{X}$ denotes the feasible solution space.
% $\mathcal{X}$ comprises all feasible solutions, and $F(\pi)$ is a $\kappa$-dimensional vector that includes $\kappa$ objective values of the solution $\pi$. 
Due to inherent conflicts among objectives, a single solution that simultaneously optimizes all objectives typically does not exist. 
Instead, Pareto optimal solutions are sought to represent different trade-offs, often guided by weight vectors, among competing objectives.
% Instead, Pareto optimal solutions are often pursued to achieve different trade-offs (i.e., weight vectors) among objectives. 
We define the Pareto properties of the solutions as follows.
% The Pareto properties of the solutions are provided in the following.

\emph{\noindent\textbf{Definition 1 (Pareto Dominance).} A solution $\pi \in \mathcal{X}$ is said to dominate another solution $\pi' \in \mathcal{X}$ (i.e., $\pi \prec \pi'$) if and only if $f_{i}(\pi) \leq f_{i}(\pi')$, $\forall i \in \left\{1, \cdots, \kappa\right\}$, and $F(\pi) \neq F(\pi')$.}

\emph{\noindent\textbf{Definition 2 (Pareto Optimality).} A solution $\pi^* \in \mathcal{X}$ is Pareto optimal if it is not dominated by any other solution. Accordingly, the Pareto set $\mathcal{P}$ is defined as all Pareto optimal solutions, i.e., $\mathcal{P} = \left\{\pi^* \in \mathcal{X} | \nexists \; \pi\in \mathcal{X}: \pi \prec \pi^* \right\}$. The Pareto front $\mathcal{F}$ is defined as images of Pareto optimal solutions in the objective space, i.e., $\mathcal{F} = \left\{F(\pi^*) | \pi^* \in \mathcal{P}\right\}$.}

\textbf{Decomposition-based Methods.} Since solving a SOCOP optimally is NP-hard, MOCOPs are significantly more intractable due to the need to identify Pareto optimal solutions, the quantity of which grows exponentially with the problem size. Therefore, MOEAs are commonly adopted to compute approximate Pareto solutions. Among them, decomposition-based MOEAs (MOEA/Ds) solve a set of subproblems (i.e., SOCOPs) derived from the original MOCOP, a foundation underlying recent DRL-based neural methods.
% by solving SOCOPs decomposed from an MOCOP, which inspires current DRL methods. 
% Particularly, the MOEA/D framework gains the solutions by solving SOCOPs, which inspires the current DRL methods. 
Specifically, the vanilla MOEA/D utilizes decomposition techniques to scalarize an MOCOP into $N$ subproblems with a set of uniformly distributed weight vectors $\{\lambda_1, \lambda_2, \cdots, \lambda_N\}$, each of which satisfies $\lambda_i = (\lambda_i^1, \cdots, \lambda_i^\kappa)^\top$, $\forall$ $\lambda_{i}^{j} \geq 0 $ and $\sum\nolimits_{j=1}^{\kappa} \lambda_{i}^{j} = 1$. 

\subsection{Subproblem Solving}
Given a subproblem ($\mathcal{G}, \lambda_i$), we formulate the solving process as a Markov Decision Process (MDP). An \emph{agent} iteratively takes the current \emph{state} as input (e.g., the instance information and the partially constructed solution), and outputs a probability distribution over the nodes to be selected next. An \emph{action} corresponds to selecting a node, either greedily or by sampling from the predicted probabilities. The \emph{transition} involves appending the selected node to the partial solution. 
We parameterize the \emph{policy} $p$ by a deep neural network $\theta$, such that the probability of constructing a complete solution $\pi$ is defined as $p_{\theta}(\pi|\mathcal{G}, \lambda_i) = \prod_{t=1}^{T} p_{\theta} (\pi_{t}|\pi_{<t},\mathcal{G}, \lambda_i)$, 
where $T$ is the total number of steps, and $\pi_{t}$ and $\pi_{<t}$ represent the selected node and partial solution at the $t$-th step, respectively.
The \emph{reward} is defined as the negation of the scalarized objective, e.g., $\mathcal{R}(\pi)=-\sum_{j=1}^{\kappa} \lambda_{i}^{j} f_{j}(\pi)$ with weighted sum decomposition. The policy network is typically optimized using the REINFORCE~\cite{williams1992simple}, which maximizes the expected reward $\mathcal{L}(\theta|\mathcal{G}, \lambda_i)=\mathbb{E}_{p_{\theta}(\pi|\mathcal{G}, \lambda_i)}\mathcal{R}(\pi)$ via the following gradient estimator:
\begin{equation}
    \label{eq:reinforce}
    % \small
    \nabla_{\theta} \mathcal{L}(\theta|\mathcal{G}, \lambda_i) = \mathbb{E}_{p_{\theta}(\pi|\mathcal{G}, \lambda_i)} [(\mathcal{R}(\pi)-b(\mathcal{G})) \nabla_{\theta}\log p_{\theta}(\pi|\mathcal{G}, \lambda_i)],
\end{equation}
where the baseline function $b(\cdot)$ reduces the gradient variance and stabilizes the training.
% over solutions to different instances, i.e., $\pi\sim p_\theta$, $\mathcal{G}\in \Tilde{\mathcal{G}}$. 
There are two primary paradigms to extend the above approach to solve MOCOPs. In the \emph{one-to-one} paradigm, methods sequentially train separate models to solve subproblems with a predefined set of weight vectors. However, they often suffer from low training efficiency and generalize poorly to subproblems with unseen weight vectors.
In contrast, the \emph{one-to-many} paradigm trains a single model to solve subproblems for arbitrary weight vectors. This is typically achieved by incorporating a subnetwork that transforms each weight vector into model parameters, thus inducing tailored policies for each subproblem.
Our POCCO follows the one-to-many paradigm and introduces two key innovations. Specifically, we design a conditional computation block to dynamically route subproblems to different neural architectures, and propose a preference-driven algorithm to train the model effectively.

% \subsection{Mixture-of-Experts and Mixture-of-Depths.}

% Recent work suggests that sparsifying a Transformer along both width and depth yields better capacity–efficiency trade-offs than targeting either axis in isolation. Mixture-of-Experts (MoE) layers widen the model: a token-level gating network $G(\mathbf{x})$ routes the input $\mathbf{x}$ to the top-k experts drawn from a pool $\{E_1,\dots,E_m\}$, and the layer output is
% $\mathrm{MoE}(\mathbf{x})=\sum_{j=1}^{m} G(\mathbf{x})_j\,E_j(\mathbf{x})$.
% Because each token is processed by only a handful of experts, MoE attains the representational power of a very wide network while containing the FLOP budget \citep{zhou2024mvmoe}.
% Complementary Mixture-of-Depths (MoD) layers thin the model: at every other Transformer block, a lightweight gate decides whether each token should execute the block or skip it, with a global capacity constraint limiting the number of tokens that can be processed \citep{raposo2024mixture}.  This dynamic depth selection reduces sequence-length–dependent computation without sacrificing expressiveness.

% Viewed together, MoE and MoD endow a Transformer with adaptive width and adaptive depth.  Tokens that are hard to model can be sent to specialized experts and traverse more layers, while easier tokens can reuse generic experts or bypass blocks entirely.  Interleaving MoE and MoD blocks therefore couples fine-grained representational diversity with conditional computation, yielding lower training loss and faster inference for large-scale language models than using either mechanism alone.

\section{Methodology}
\label{sec:method}

\subsection{Overview}
% POCCO is a learning-based approach designed to learn a set of policies for constructing solutions to SOCOPs $\{(\mathcal{G},\lambda_{i})\}_{i=1}^{N}$ in sync, which are decomposed from an MOCOP $\mathcal{G}$. During training, each policy is allowed to specialize on a subset of subproblems instead of learning a generalized policy capable of all subproblem which is generally lack of diversity. This essentially results in a portfolio of policies, which are known to be highly effective for solving multi-task problems\cite{wang2024scaling}. Our pursuit of diversity is fundamentally a means to enhance exploration and consequently solution quality. An intuitive approach to realize this diversity is to activate different network parameters for obtain different policies. On the other hand, POCCO expects to uniformly learn these policies avoiding of some policies dominating the training process which is often caused by imbalance reward signals. Therefore, POCCO exploits preference signals to train the deep model by maximizing the likelihood of wining solutions and minimizing the likelihood of losing method. To achieve this, we construct a set of wining-losing solution pairs $\{\pi_w^i, \pi_l^i\}_{i=1}^M$ for a SOCOP instance $(\mathcal{G},\lambda_{i})$. 
POCCO is a learning-based framework that trains a portfolio of policies to solve a set of scalarized subproblems $\{(\mathcal{G},\lambda_i)\}_{i=1}^{N}$, obtained by decomposing an MOCOP instance. Instead of forcing a single policy to handle all subproblems, which often yields bland and suboptimal behavior, POCCO promotes specialization: each policy is encouraged to focus on a subset of subproblems, yielding a diverse policy ensemble.
Such diversity is known to enhance multi-task optimization \cite{wang2024scaling} by expanding the exploration space and ultimately improving solution quality.
% that is known to benefit multi-task optimization \cite{wang2024scaling}.  Diversity here serves a critical purpose: it broadens exploration and, in turn, improves solution quality.
Technically, we achieve this diversity by activating different subsets of model parameters through a CCO block, enabling distinct computational paths to emerge for different subproblems. 
Moreover, POCCO should encourage each policy to thoroughly explore the combinatorial solution space during training for reducing suboptimality.
% as insufficient exploration is a common failure mode that leads to suboptimal performance on subproblems. 
To achieve this, we replace raw rewards with preference signals. For each subproblem $(\mathcal{G},\lambda_i)$, we construct a set of winning–losing solution pairs $\{(\pi^{w,j}, \pi^{l,j})\}_{j=1}^{K}$. Training then proceeds by maximizing the likelihood of the winning solutions while minimizing that of the losing ones, following a BT-style objective. 
% This preference-driven training \colorb{drives the policy to focus on informative comparisons rather than absolute outcomes, facilitating more efficient learning.}
This preference-driven training encourages the learned policy to explore the most promising regions of the search space, leading to more efficient convergence toward higher-quality solutions.
% As a result, the policy is steered toward the most promising regions of the search space and converges to higher-quality solutions more efficiently.
% \colorr{This preference-driven training encourages the learned policy to explore preferred solutions effectively, thereby enhancing solving performance.}
% equalizes learning pressure across policies, preserves diversity, and yields a robust ensemble well-suited to the entire MOCOP spectrum.
% At the same time, POCCO must ensure that no single policy dominates training, which is a common failure mode when reward magnitudes differ sharply across subproblems. To balance learning,
% Our POCCO is a generic framework that can be integrated with existing neural MOCO methods. We implement it with the two, CNH and WE-CA, to further enhance their performance in solving MOCOPs. Both their models consist of an encoder that creates an embedding integrating the weight-instance interaction, and a decoder that generates multiple solutions for an instance based on the the embedding. To generate solution quikly and effectively, we only insert the conditional computation block into the decoder, allowing us to generate multiple diverse solutions for an instance with only a single pass through the computationally expensive encoder.
Notably, POCCO is a generic, plug-and-play framework that can seamlessly involve different neural solvers for MOCOPs. 
% We demonstrate this by augmenting two state-of-the-art methods, CNH and WE-CA. In both methods, an encoder produces a joint weight–instance embedding, and a decoder generates candidate solutions conditioned on that embedding. To boost efficiency and diversity, we insert POCCO’s conditional-computation block only in the decoder. This design lets us draw multiple diverse solutions from a single, costly encoder pass, thereby improving both speed and overall solution quality.
We demonstrate this by augmenting two SOTA methods, CNH \cite{fan2024conditional} and WE-CA \cite{chen2025rethinking}, in Section \ref{sec:experiments}. 

% In both, the encoder generates joint node embeddings $\{h_i\}_{i=0}^{n}$ that capture the interaction between the instance $\mathcal{G}$ and the weight vector $\lambda_i$, and the decoder produces candidate solutions conditioned on these embeddings. To maintain efficiency and enhance diversity, we integrate the proposed CCO block solely into the decoder. This design enables the generation of multiple diverse solutions from a single, computationally expensive encoder pass, delivering a favorable trade-off between empirical performance and computational cost.

\subsection{Conditional Computational Block}
Most approaches employ a Transformer-based architecture, where the encoder generates joint node embeddings $\{h_i\}_{i=0}^{n}$ that capture the interaction between the instance $\mathcal{G}$ and the weight vector $\lambda_i$, and the decoder produces candidate solutions conditioned on these embeddings. We propose a CCO block to increase the model capacity and promote policy diversity across subproblems. To maintain efficiency, we integrate the CCO block solely into the decoder of the backbone model. This design enables the generation of multiple diverse solutions through a single, computationally expensive encoder pass, offering a favorable trade-off between empirical performance and computational cost.

\begin{figure}[!t]
    \centering
    \vspace{-2mm}
    \includegraphics[width=0.95\textwidth]{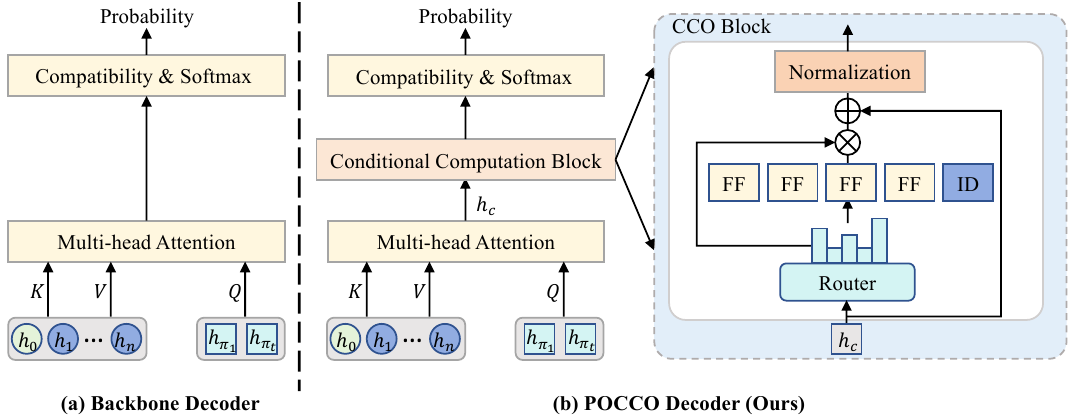}
    \caption{Decoder structures of backbone and POCCO. Given an MOCOP instance $\mathcal{G}$ with $n+1$ nodes (e.g., $n$ customers and a depot, if applicable) and a weight vector $\lambda_i$, POCCO encodes their raw features into joint node embeddings $\{h_i\}_{i=0}^{n}$ using an encoder. At each decoding step $t$, the decoder forms a query $Q$ from the embeddings of the first and last selected nodes $(\pi_1,\pi_t)$, and computes the key $K$ and value $V$ via linear projections of $\{h_i\}_{i=0}^{n}$. The MHA layer processes $Q$, $K$, and $V$ to produce a context vector $h_c$, which is refined by the CCO block. The refined context is passed through a compatibility layer followed by a Softmax to compute the node selection probabilities. More details about the forward pass can be found in Appendix \ref{app:b}.}
    \label{fig:cco}
    \vspace{-3mm}
\end{figure}

As illustrated in Fig. \ref{fig:cco}, the CCO block comprises multiple FF experts and a single ID expert. 
We insert this block between the multi-head attention (MHA) layer and the compatibility layer in the decoder.
Given a batch of MHA outputs $\{h_c^b\}_{b=1}^{B}$, the CCO block dynamically routes each context vector $h_c^b$ from the corresponding subproblem through either the FF or ID experts, forming distinct computation paths that function as different policies. The ID expert allows the model to bypass the FF computation, promoting architectural sparsity and specialization \cite{han2025slim}. Consequently, the CCO block facilitates the learning of dedicated, weight-specific policies tailored to individual subproblems. 

% \noindent\textbf{The CCO block.}
Formally, a CCO block consists of: 1) $m$ FF experts $\{E_1, E_2, \dots, E_m\}$ with independent trainable parameters; 2) a parameter-free identity expert $E_{m+1}$; 3) a router, implemented as a gating network $G$ parameterized by $W_G$, which determines how the inputs $\{h_c^{b}\}_{b=1}^{B}$ are routed to the experts; and 4) a skip connection followed by an instance normalization (IN) layer.
Given a single context vector $h_c^b$, let $G(h_c^b) \in \mathbb{R}^{m+1}$ denote the output of the gating network, which represents the expert selection probabilities, and let $E_j(h_c^b)$ denote the output of the $j$‑th expert. The output of the CCO block is:
\begin{equation}
\operatorname{CCO}(h_c^b)=\operatorname{IN} \left(\sum_{j=1}^{m+1} G(h_c^b)_j E_j(h_c^b)+h_c^b\right).
\label{eq:cco}
\end{equation}
The sparse vector $G(h_c^b)$ activates only a small subset of experts, either parameterized FF experts or the parameter‑free ID expert, thereby enabling diverse computation paths while reducing computational overhead. A typical implementation uses a $\operatorname{Top}k$ operator that retains the $k$ largest logits and masks the rest with $-\infty$.  In this case, the gating network output is: $G(h_c^b)=\operatorname{Softmax} \bigl(\operatorname{Top}k(h_c^b \cdot W_G)\bigr)$.
% \begin{equation}
% G(h_c^b)=\operatorname{Softmax} \bigl(\operatorname{Top}k(h_c^b W_G)\bigr).
% \end{equation}

Our proposed CCO block aligns with the principles of recent advances in efficiently scaling Transformer-based models along both width \cite{shazeer2017} and depth \cite{raposo2024mixture}. 
In specific, it combines a mixture‑of‑experts (MoE) layer, implemented using multiple FF experts to widen the network, with a mixture‑of‑depths (MoD) layer, realized through an ID expert that allows inputs to skip computation. Within the CCO block, each subproblem is adaptively routed to only a small subset of experts, granting the model the expressiveness of a significantly wider network while preserving computational efficiency. 
As demonstrated in Section \ref{sec:experiments}, this joint design achieves a better capacity–efficiency trade-off than scaling either dimension in isolation.
% This MoDE design offers adaptive width and depth: complex sub‑problems are handled by specialized experts, whereas simpler ones bypass unnecessary computations, achieving both accuracy and efficiency.
% This design yields better capacity–efficiency trade-offs than targeting either axis in isolation as demonstrated in Section \ref{sec:experiments}.

% Recent work suggests that sparsifying a Transformer along both width and depth yields better capacity–efficiency trade-offs than targeting either axis in isolation~\cite{zhou2024mvmoe,raposo2024mixture}. Building on this insight, our proposed CCO combines a Mixture‑of‑Experts (MoE) layer, implemented as multiple feed‑forward experts that widen the network, with a Mixture‑of‑Depths (MoD) component realized through an ID expert that allows tokens to skip computation. Within the CCO block, each sub‑problem is routed to only a small set of experts, giving the model the expressiveness of a much wider network while keeping the FLOP budget under control. This MoDE design offers adaptive width and depth: complex sub‑problems are handled by specialized experts, whereas simpler ones bypass unnecessary computations, achieving both accuracy and efficiency.

\subsection{Preference-driven MOCO}
\begin{figure}[!t]
    \vspace{-2mm}
    \centering
    \includegraphics[width=0.98\textwidth]{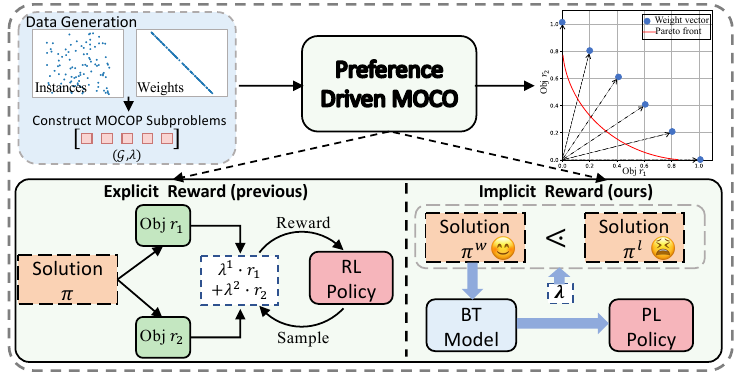}
    % \vspace{-3mm}
    \caption{An overview of preference-driven MOCO. 
    % \colorr{Generated subproblems consisting of instances and weight vectors are fed into POCCO, which then returns solutions approximating the Pareto front. 
    Unlike prior DRL methods that explicitly learn from scalarized rewards, our approach converts relative preferences into a BT likelihood, providing an implicit reward signal to optimize the PL policy.}
    \vspace{-2mm}
    \label{fig:po}
\end{figure}

% To remove the reward‑scale imbalance that hampers REINFORCE‑style learning, we optimize \emph{relative preferences} instead of absolute objective values. 
To mitigate the exploration inefficiencies inherent in REINFORCE algorithms, we optimize \emph{relative preferences} \cite{meng2024simpo} instead of absolute objective values.
We summarize our preference-driven MOCO in Appendix~\ref{appendix:alg} and outline the complete pipeline in Fig. \ref{fig:po}, which proceeds in three steps as follows.
% Fig. \ref{fig:po} outlines the complete pipeline of our preference-driven MOCO, which proceeds in three steps as follows.

\noindent\textbf{Generating preference pairs.}
For each scalarized subproblem $(\mathcal{G}, \lambda_i)$ in the training batch, the policy $p_{\theta}$ samples two candidate solutions, $\pi^{w}$ and $\pi^{l}$. We denote $\pi^{w} \lessdot \pi^{l}$ if $\pi^{w}$ is preferred over $\pi^{l}$, as determined by the ordering of their scalarized objective values. This evaluator ranks the two solutions and designates the better one as the winning solution $\pi^{w}$ and the other as the losing solution $\pi^{l}$. A binary preference label $y$ is then assigned, where $y = 1$ if $\pi^{w} \lessdot \pi^{l}$, and $y = 0$ otherwise. This label serves as the supervision signal required for the preference-driven MOCO.

\noindent\textbf{Defining an implicit reward.}
% Directly using raw objective values would re‑introduce the very scale mismatch we seek to avoid. Instead, we follow \cite{meng2024simpo} and treat the average log‑likelihood of a solution as a implicit reward that is inherently normalized by sequence length:
Distinct from DRL training paradigms that rely on raw objective values, POCCO treats the average log‑likelihood of a solution as an implicit reward $f_{\theta}$, directly relating preferences between solutions to their policy probabilities. This reward is inherently normalized by sequence length $|\pi|$, thereby mitigating length bias between winning and losing solution pairs.
\begin{equation}
\label{eq:simpo_reward}
    f_\theta(\pi|\mathcal{G},\lambda_i)= \frac{1}{|\pi|} \log\, p_{\theta}(\pi|\mathcal{G}, \lambda_i) = \frac{1}{|\pi|}\sum_{t=1}^{|\pi|}\log\, p_\theta(\pi_t| \pi_{<t}, \mathcal{G},\lambda_i).
\end{equation}
% where $|\pi|$ is the sequence length and $p_{\theta}$ is the autoregressive policy network. 
% Because log‑likelihoods are on a consistent numerical scale for all subproblems, they sidestep the reward‑magnitude disparity across different weight vectors $\lambda$.

\textbf{Learning from pairwise comparisons.}
We formulate preference learning (PL) as a probabilistic binary classification problem through the BT model. Specifically, the BT model is a pairwise preference framework that uses a function $g_\theta(\cdot)$ to map reward differences into preference probabilities.
% \colorr{The BT model is a paired preference model which can be used to relate reward differences to preferences. In BT model, a function $g_\theta(\cdot)$ maps the difference between reward into preference probability.}
It assigns each solution a strength proportional to its implicit reward (defined in Eq. (\ref{eq:simpo_reward})) and predicts the probability $g_{\theta}$ that the winning solution outranks the losing one:
\begin{equation}
g_{\theta}\!\bigl(\pi^{w}\lessdot\pi^{\ell}\mid\mathcal{G},\lambda_i \bigr)=
\sigma \bigl( \beta\,[\,f_{\theta}(\pi^{w}\mid\mathcal{G},\lambda_i) - f_{\theta}(\pi^{\ell}\mid\mathcal{G},\lambda_i)\,]\bigr),
\end{equation}
where $\sigma(\cdot)$ is the sigmoid function, and $\beta>0$ is a fixed temperature that controls the sharpness with which the model distinguishes between unequal rewards.
% We maximize the log‑likelihood of \colorb{(correct -> ground truth?)} preferences, yielding the following loss function:
We maximize the likelihood of the collected preferences, yielding the following loss function:
\begin{equation}
\label{eq:po_loss}
 \mathcal{L}(\theta|p_\theta, \mathcal{G},\lambda_i, \pi^w, \pi^l) = - y\,\log\, \sigma \bigl( \beta [\frac{\log\, p_\theta(\pi^w|\mathcal{G},\lambda_i)}{|\pi^w|}-\frac{\log\, p_\theta(\pi^l|\mathcal{G},\lambda_i)}{|\pi^l|}\bigr]).
\end{equation}

% \emph{Scale invariance}: Equation~\ref{eq:po_loss} depends only on the \emph{difference} of two log‑likelihoods, so every preference pair contributes gradients of comparable magnitude, even when the underlying objectives differ by orders of magnitude.
% \emph{Exploration}: Because the policy’s own likelihood scores serve as the reward signal, the model is automatically encouraged to explore solutions it currently underestimates.
% \emph{Simplicity}: No separate value network or reward normalizer is required; the policy $g_{\theta}$ alone suffices.
In practice, we collect multiple $(\pi^{w},\pi^{\ell})$ pairs per update, sum their losses, and backpropagate through $p_{\theta}$. 
% This results in a stable, reward-agnostic training procedure that \colorb{remains well-balanced across all subproblems} in the MOCOP setting.
By maximizing the log-likelihood of $g_{\theta}\!\bigl(\pi^{w}\lessdot\pi^{\ell}\mid\mathcal{G},\lambda_i \bigr)$, the model is encouraged to assign higher probabilities to preferred solutions $\pi^w$ over less preferred ones $\pi^l$.
% We include the preference-driven MOCO algorithm in Algorithm~\ref{alg:po}.

% \noindent\textbf{Importance Sampling.} The backbone models (i.e., CNH and WE-CA) are designed to train a single unified policy across a range of problem sizes. In their original implementations, each training batch samples a problem size uniformly from a predefined interval. However, this uniform sampling strategy introduces a bias: smaller instances are overrepresented during training, potentially leading the model to overfit on easy, small-scale problems while underperforming on larger, more challenging instances.
% To address this imbalance, we propose an importance sampling strategy that assigns higher sampling probabilities to larger problem sizes, reflecting their greater learning importance and difficulty. Specifically, we use the problem size as a proxy for importance, and define the sampling weights linearly based on size. This strategy ensures that larger problem instances are sampled more frequently during training, encouraging the model to better generalize across the full size spectrum and improving performance on more complex cases.

\vspace{-2mm}
% \newpage
\section{Experiments}
\label{sec:experiments}

\subsection{Experimental settings}
\textbf{Training.} 
% Revisions
% problems
We conduct extensive experiments to evaluate the effectiveness of the proposed POCCO across various MOCOPs, including multi-objective traveling salesman problem (MOTSP)~\cite{lust2010multiobjective}, multi-objective capacitated vehicle routing problem (MOCVRP)~\cite{zajac2021objectives}, and multi-objective knapsack problem (MOKP)~\cite{ishibuchi2014behavior}. In MOTSP, the goal is to find a tour that visits all nodes exactly once, while minimizing multiple total path lengths, each computed based on a distinct set of coordinates associated with a specific objective. Regarding MOCVRP, a fleet of vehicles with limited capacity must serve all customer nodes and return to the depot, with the objectives of minimizing the total travel distance and the length of the longest individual route. As for MOKP, the problem involves selecting a subset of items, each with a weight and multiple objective-specific values. The objective is to maximize all objective values simultaneously, while ensuring the total weight stays within the knapsack capacity. In this work, we consider three commonly used problem sizes: $n = 20/50/100$ for MOTSP and MOCVRP, and $n = 50/100/200$ for MOKP.

\noindent\textbf{Hyperparameters.}
We implement POCCO on top of two SOTA neural MOCO methods, CNH~\cite{fan2024conditional} and WE-CA~\cite{chen2025rethinking}, resulting in POCCO-C and POCCO-W, respectively. Most hyperparameters are aligned with those used in the original CNH and WE-CA implementations. Both models are trained for 200 epochs, with each epoch processing 100,000 randomly sampled instances and a batch size of $B = 64$. We use the Adam optimizer~\cite{kingma2014adam} with a learning rate of $3 \times 10^{-4}$ and a weight decay of $10^{-6}$. The $N$ weight vectors used for decomposition are generated following~\cite{das1996normal}, with $N = 101$ for $\kappa = 2$ and $N = 105$ for $\kappa = 3$.

\begin{table}[!t]
  \centering
  \caption{Performance on BiTSP and MOCVRP Instances}
  \label{tab:bitsp}
  \begin{small}
    \renewcommand\arraystretch{0.5}
    % BiTSP block
    \resizebox{0.98\textwidth}{!}{
      \begin{tabular}{lccc ccc ccc}
        \toprule
        & \multicolumn{3}{c}{Bi-TSP20}
        & \multicolumn{3}{c}{Bi-TSP50}
        & \multicolumn{3}{c}{Bi-TSP100} \\
        \cmidrule(lr){2-4}\cmidrule(lr){5-7}\cmidrule(lr){8-10}
        Method       & HV      & Gap      & Time   & HV      & Gap      & Time   & HV      & Gap       & Time   \\
        \midrule
        WS-LKH     & 0.6270 & 0.00\%   & 10m   & 0.6415 & 0.05\%   & 1.8h  & \textbf{0.7090} & -0.17\%   & 6h    \\
        \midrule
      MOEA/D     & 0.6241 & 0.46\%   & 1.7h  & 0.6316 & 1.59\%   & 1.8h  & 0.6899 & 2.53\%    & 2.2h  \\
      NSGA-II    & 0.6258 & 0.19\%   & 6.0h  & 0.6120 & 4.64\%   & 6.1h  & 0.6692 & 5.45\%    & 6.9h  \\
      MOGLS      & \textbf{0.6279} & -0.14\%  & 1.6h  & 0.6330 & 1.37\%   & 3.7h  & 0.6854 & 3.16\%    & 11h   \\
      PPLS/D-C   & 0.6256 & 0.22\%   & 26m   & 0.6282 & 2.12\%   & 2.8h  & 0.6844 & 3.31\%    & 11h   \\
      \midrule
      DRL-MOA    & 0.6257 & 0.21\%   & 6s    & 0.6360 & 0.90\%   & 9s    & 0.6970 & 1.53\%    & 16s   \\
      MDRL       & 0.6271 & -0.02\%  & 5s    & 0.6364 & 0.84\%   & 8s    & 0.6969 & 1.54\%    & 14s   \\
      EMNH       & 0.6271 & -0.02\%  & 5s    & 0.6364 & 0.84\%   & 8s    & 0.6969 & 1.54\%    & 15s   \\
      PMOCO      & 0.6259 & 0.18\%   & 6s    & 0.6351 & 1.04\%   & 12s    & 0.6957 & 1.71\%    & 26s   \\
      \midrule
      CNH        & 0.6270 & 0.00\%   & 13s   & 0.6387 & 0.48\%   & 16s   & 0.7019 & 0.83\%    & 33s   \\
      \textbf{POCCO-C}    & \underline{0.6275} & -0.08\%  & 14s   & 0.6409 & 0.14\%   & 20s   & 0.7047 & 0.44\%    & 42s   \\
      \midrule
      WE-CA      & 0.6270 & 0.00\%   & 6s    & 0.6392 & 0.41\%   & 9s    & 0.7034 & 0.62\%    & 18s   \\
      \textbf{POCCO-W}    & \underline{0.6275} & -0.08\%  & 7s    & 0.6411 & 0.11\%   & 14s   & 0.7055 & 0.32\%    & 36s   \\
      \midrule
      MDRL-Aug   & 0.6271 & -0.02\%  & 47s   & 0.6408 & 0.16\%   & 1.8m  & 0.7022 & 0.79\%    & 5.4m  \\
      EMNH-Aug   & 0.6271 & -0.02\%  & 46s   & 0.6408 & 0.16\%   & 1.8m  & 0.7023 & 0.78\%    & 5.4m  \\
      PMOCO-Aug  & 0.6270 & 0.00\%   & 39s   & 0.6395 & 0.36\%   & 1.7m  & 0.7016 & 0.88\%    & 5.8m  \\
      \midrule
      CNH-Aug    & 0.6271 & -0.02\%  & 1.3m  & 0.6410 & 0.12\%   & 3.9m  & 0.7054 & 0.34\%    & 12m   \\
      \textbf{POCCO-C-Aug} & 0.6270  & 0.00\%   & 2.2m  & \underline{0.6416} & 0.03\%   & 4.0m  & 0.7071 & 0.10\%    & 14m   \\
      \midrule
      WE-CA-Aug  & 0.6271 & -0.02\%  & 1.3m  & 0.6413 & 0.08\%   & 3.6m  & 0.7066 & 0.17\%    & 12m   \\
      \textbf{POCCO-W-Aug}& 0.6270 & 0.00\%   & 2.2m  & \textbf{0.6418} & 0.00\%   & 4.0m  & \underline{0.7078} & 0.00\%    & 14m   \\
        % \bottomrule
      \end{tabular}}
    \vspace{-0.1ex}
    % MOCVRP block
    \resizebox{0.98\textwidth}{!}{
      \begin{tabular}{lccc ccc ccc}
        \toprule
          & \multicolumn{3}{c}{MOCVRP20}
          & \multicolumn{3}{c}{MOCVRP50}
          & \multicolumn{3}{c}{MOCVRP100} \\
          \cmidrule(lr){2-4}\cmidrule(lr){5-7}\cmidrule(lr){8-10}
          Method & HV     & Gap     & Time    & HV     & Gap      & Time    & HV     & Gap       & Time   \\
        \midrule
        MOEA/D      & 0.4255 & 1.07\%   & 2.3h   & 0.4000 & 2.63\%   & 2.9h   & 0.3953 & 3.33\%    & 5.0h   \\
      NSGA-II     & 0.4275 & 0.60\%   & 6.4h   & 0.3896 & 5.16\%   & 8.8h   & 0.3620 & 11.47\%   & 9.4h   \\
      MOGLS       & 0.4278 & 0.53\%   & 9.0h   & 0.3984 & 3.02\%   & 20h    & 0.3875 & 5.23\%    & 72h    \\
      PPLS/D-C    & 0.4287 & 0.33\%   & 1.6h   & 0.4007 & 2.46\%   & 9.7h   & 0.3946 & 3.50\%    & 38h    \\
      \midrule
      DRL-MOA     & 0.4287 & 0.33\%   & 8s     & 0.4076 & 0.78\%   & 12s    & 0.4055 & 0.83\%    & 21s    \\
      MDRL        & 0.4291 & 0.23\%   & 6s     & 0.4082 & 0.63\%   & 13s    & 0.4056 & 0.81\%    & 22s    \\
      EMNH        & 0.4299 & 0.05\%   & 7s     & 0.4098 & 0.24\%   & 12s    & 0.4072 & 0.42\%    & 22s    \\
      PMOCO       & 0.4267 & 0.79\%   & 6s     & 0.4036 & 1.75\%   & 12s    & 0.3913 & 4.30\%    & 22s    \\
      \midrule
      CNH         & 0.4287 & 0.33\%   & 11s    & 0.4087 & 0.51\%   & 15s    & 0.4065 & 0.59\%    & 25s    \\
      \textbf{POCCO-C}     & 0.4294 & 0.16\%   & 16s    & 0.4101 & 0.17\%   & 25s    & 0.4079 & 0.24\%    & 53s    \\
      \midrule
      WE-CA       & 0.4290 & 0.26\%   & 7s     & 0.4089 & 0.46\%   & 10s    & 0.4068 & 0.51\%    & 21s    \\
      \textbf{POCCO-W}     & 0.4294 & 0.16\%   & 8s     & 0.4102 & 0.15\%   & 17s    & 0.4084 & 0.12\%    & 46s    \\
      \midrule
      MDRL-Aug    & 0.4294 & 0.16\%   & 12s    & 0.4092 & 0.39\%   & 36s    & 0.4072 & 0.42\%    & 2.8m   \\
      EMNH-Aug    & \textbf{0.4302} & -0.02\%  & 12s    & \underline{0.4106} & 0.05\%   & 35s    & 0.4079 & 0.24\%    & 2.8m   \\
      PMOCO-Aug   & 0.4294 & 0.16\%   & 14s    & 0.4080 & 0.68\%   & 42s    & 0.3969 & 2.93\%    & 2.0m    \\
      \midrule
      CNH-Aug     & 0.4299 & 0.05\%   & 21s    & 0.4101 & 0.17\%   & 45s    & 0.4077 & 0.29\%    & 1.9m   \\
      \textbf{POCCO-C-Aug} & \textbf{0.4302} & -0.02\%  & 31s    & \textbf{0.4108} & 0.00\%   & 1.4m   & \underline{0.4086} & 0.07\%    & 2.4m   \\
      \midrule
      WE-CA-Aug   & 0.4300 & 0.02\%   & 15s    & 0.4103 & 0.12\%   & 36s    & 0.4081 & 0.20\%    & 1.8m   \\
      \textbf{POCCO-W-Aug} & \underline{0.4301} & 0.00\%   & 24s    & \textbf{0.4108} & 0.00\%   & 1.2m   & \textbf{0.4089} & 0.00\%    & 2.3m   \\
        \bottomrule
       \vspace{-4mm}
      \end{tabular}}
  \end{small}
\end{table}
% \renewcommand\thefootnote{\fnsymbol{footnote}}
% \footnotetext[2]{Note that POCCO with instance augmentation yields lower HV values compared to its non-augmented counterpart. This is because decomposition-based methods focus on optimizing individual subproblems rather than ensuring overall solution diversity. While augmentation improves solution quality for specific subproblems, it may reduce the number of non-dominated solutions, resulting in a smaller HV.}
% \renewcommand\thefootnote{\arabic{footnote}}

\noindent\textbf{Baselines.}
% Revisions for Baselines
We compare POCCO with a broad range of baseline methods across three categories, all employing weighted-sum (WS) scalarization to ensure fair comparison: (1) Single-model neural MOCO approaches: This includes \textbf{PMOCO}~\cite{lin2022pareto},  and recent SOTA methods \textbf{CNH}~\cite{fan2024conditional}, and \textbf{WE-CA}~\cite{chen2025rethinking}. Both CNH and WE-CA are unified models trained across problem size $n \in \{20, 21, \cdots, 100\}$ (except $n\in\{50, 51, \cdots, 200\}$ for Bi-KP) (2) Multi-model neural MOCO approaches: This category covers methods like \textbf{DRL-MOA}~\cite{likaiwen2020deep}, \textbf{MDRL}~\cite{zhang2022meta}, and \textbf{EMNH}~\cite{chen2024efficient}. 
Specifically, DRL-MOA trains a separate POMO model for each of the $N$ subproblems, with the first model trained for 200 epochs and the rest fine-tuned for 5 epochs each using parameter transfer. MDRL and EMNH both initialize from a shared pretrained meta-model and fine-tune $N$ subproblem-specific models using the same network structure and training settings as in~\cite{chen2024efficient}. 
(3) Non-learnable approaches, including classical MOEAs and other problem-specific heuristics: \textbf{MOEA/D}~\cite{zhang2007moea} and \textbf{NSGA-II}~\cite{deb2002fast}, each run for 4,000 iterations, serve as representative decomposition-based and dominance-based MOEAs, respectively. MOCOP-specific MOEAs such as \textbf{MOGLS}~\cite{jaszkiewicz2002genetic}, configured with 4,000 iterations and 100 local search steps per iteration, and \textbf{PPLS/D-C}~\cite{shi2022improving}, run for 200 iterations, are also considered. These methods use 2-opt heuristics for MOTSP and MOCVRP, and a greedy transformation heuristic~\cite{ishibuchi2014behavior} for MOKP. Finally, \textbf{WS-LKH} and \textbf{WS-DP} combine weighted-sum scalarization with powerful solvers, with LKH~\cite{helsgaun2000effective, tinos2018efficient} used for MOTSP and dynamic programming applied to MOKP.

\noindent\textbf{Inference.}
% Revisions for Metrics
We evaluate all methods using three metrics: average hypervolume (HV)~\cite{while2006faster}, average gap, and total runtime per instance set. HV is a widely used indicator in multi-objective optimization that reflects both the convergence and diversity of the solution set. A higher HV indicates better performance. To ensure consistency, HV values are normalized to the range $[0, 1]$ using the same reference point for all methods. The gap is defined as the relative difference between a method’s HV and the HV of POCCO-W. Methods with the “-Aug” suffix apply instance augmentation~\cite{lin2022pareto} to further improve performance. To evaluate statistical significance, we use the Wilcoxon rank-sum test~\cite{wilcoxon1992individual} at a 1\% significance level. The best results and others that are not significantly worse are marked in bold, while the second-best and statistically similar results are underlined. All experiments are implemented in Python and conducted on a machine with NVIDIA Ampere A100-80GB GPUs and an AMD EPYC 7742 CPU. 
The code and dataset are publicly released for reproducibility.\footnote{https://github.com/mingfan321/POCCO}
% The code and dataset will be released publicly upon acceptance~\footnote{https://github.com/mingfan321/POCCO}.

\begin{table}[!t]
\centering
\caption{Performance comparison on Bi-KP and Tri-TSP Instances}
\label{tab:tritsp}
\begin{small}
\renewcommand\arraystretch{0.5}
\resizebox{0.95\textwidth}{!}{%
\begin{tabular}{l ccc ccc ccc}
\toprule
\multirow{2}{*}{Method}
  & \multicolumn{3}{c}{Bi-KP50} & \multicolumn{3}{c}{Bi-KP100} & \multicolumn{3}{c}{Bi-KP200} \\
\cmidrule(lr){2-4} \cmidrule(lr){5-7} \cmidrule(lr){8-10}
  & HV & Gap & Time & HV & Gap & Time & HV & Gap & Time \\
\midrule
      WS-DP      & \underline{0.3561}  & 0.03\%   & 22m    & 0.4532  & 0.04\%   & 2h     & 0.3601  & 0.06\%   & 5.8h   \\
      \midrule
      MOEA/D     & 0.3540  & 0.62\%   & 1.6h   & 0.4508  & 0.57\%   & 1.7h   & 0.3581  & 0.61\%   & 1.8h   \\
      NSGA-II    & 0.3547  & 0.42\%   & 7.8h   & 0.4520  & 0.31\%   & 8.0h   & 0.3590  & 0.36\%   & 8.4h   \\
      MOGLS      & 0.3540  & 0.62\%   & 5.8h   & 0.4510  & 0.53\%   & 10h    & 0.3582  & 0.58\%   & 18h    \\
      PPLS/D-C   & 0.3528  & 0.95\%   & 18m    & 0.4480  & 1.19\%   & 47m    & 0.3541  & 1.72\%   & 1.5h   \\
      \midrule
      DRL-MOA    & 0.3559  & 0.08\%   & 8s     & 0.4531  & 0.07\%   & 15s    & 0.3601  & 0.06\%   & 32s    \\
      MDRL       & 0.3530  & 0.90\%   & 7s     & 0.4532  & 0.04\%   & 18s    & 0.3601  & 0.06\%   & 35s    \\
      EMNH       & \underline{0.3561}  & 0.03\%   & 7s     & \textbf{0.4535}  & -0.02\%  & 17s    & \textbf{0.3603}  & 0.00\%   & 48s    \\
      PMOCO      & 0.3552  & 0.28\%   & 8s     & 0.4523  & 0.24\%   & 22s    & 0.3595  & 0.22\%   & 50s    \\
      \midrule
      CNH        & 0.3556  & 0.17\%   & 16s    & 0.4527  & 0.15\%   & 23s    & 0.3598  & 0.14\%   & 55s    \\
      \textbf{POCCO-C}    & 0.3560  & 0.06\%   & 20s    & \textbf{0.4535}  & -0.02\%  & 36s    & \textbf{0.3603}  & 0.00\%   & 1.4m   \\
      \midrule
      WE-CA      & 0.3558  & 0.11\%   & 8s     & 0.4531  & 0.07\%   & 16s    & \underline{0.3602}  & 0.03\%   & 50s    \\
      \textbf{POCCO-W}    & \textbf{0.3562}  & 0.00\%   & 11s    & \underline{0.4534}  & 0.00\%   & 26s    & \textbf{0.3603}  & 0.00\%   & 1.3m   \\
% \bottomrule
      \end{tabular}}%

    \vspace{-0.1ex}
    % MOCVRP block
    \resizebox{0.95\textwidth}{!}{%
      \begin{tabular}{lccc ccc ccc}
        \toprule
    & \multicolumn{3}{c}{Tri-TSP20}
      & \multicolumn{3}{c}{Tri-TSP50}
      & \multicolumn{3}{c}{Tri-TSP100} \\
      \cmidrule(lr){2-4}\cmidrule(lr){5-7}\cmidrule(lr){8-10}
      Method    & HV      & Gap       & Time   & HV      & Gap       & Time   & HV      & Gap        & Time   \\
\midrule
    WS-LKH     & \textbf{0.4712} & 0.00\%   & 12m   & \textbf{0.4440} & -0.07\%  & 1.9h  & \textbf{0.5076} & -0.55\%   & 6.6h   \\
    \midrule
      MOEA/D     & 0.4702 & 0.21\%   & 1.9h  & 0.4314 & 2.77\%   & 2.2h  & 0.4511 & 10.64\%   & 2.4h   \\
      NSGA-II    & 0.4238 & 10.06\%  & 7.1h  & 0.2858 & 35.59\%  & 7.5h  & 0.2824 & 44.06\%   & 9.0h   \\
      MOGLS      & 0.4701 & 0.23\%   & 1.5h  & 0.4211 & 5.09\%   & 4.1h  & 0.4254 & 15.73\%   & 13h    \\
      PPLS/D-C   & 0.4698 & 0.30\%   & 1.4h  & 0.4174 & 5.93\%   & 3.9h  & 0.4376 & 13.31\%   & 14h    \\
      \midrule
      DRL-MOA    & 0.4699 & 0.28\%   & 6s    & 0.4303 & 3.02\%   & 9s    & 0.4806 & 4.79\%    & 18s    \\
      MDRL       & 0.4699 & 0.28\%   & 5s    & 0.4317 & 2.70\%   & 10s   & 0.4852 & 3.88\%    & 17s    \\
      EMNH       & 0.4699 & 0.28\%   & 5s    & 0.4324 & 2.55\%   & 10s   & 0.4866 & 3.61\%    & 17s    \\
      PMOCO      & 0.4693 & 0.40\%   & 5s    & 0.4315 & 2.75\%   & 12s   & 0.4858 & 3.76\%    & 33s    \\
      \midrule
      CNH        & 0.4698 & 0.30\%   & 10s   & 0.4358 & 1.78\%   & 14s   & 0.4931 & 2.32\%    & 25s    \\
      \textbf{POCCO-C}    & 0.4704 & 0.17\%   & 18s   & 0.4393 & 0.99\%   & 17s   & 0.4985 & 1.25\%    & 28s    \\
      \midrule
      WE-CA      & 0.4707 & 0.11\%   & 5s    & 0.4389 & 1.08\%   & 8s    & 0.4975 & 1.45\%    & 17s    \\
      \textbf{POCCO-W}    & \underline{0.4710} & 0.04\%   & 6s    & 0.4397 & 0.90\%   & 13s   & 0.4985 & 1.25\%    & 23s    \\
      \midrule
      MDRL-Aug   & \textbf{0.4712} & 0.00\%   & 4.2m  & 0.4408 & 0.65\%   & 25m   & 0.4958 & 1.78\%    & 1.6h   \\
      EMNH-Aug   & \textbf{0.4712} & 0.00\%   & 4.2m  & 0.4418 & 0.43\%   & 25m   & 0.4973 & 1.49\%    & 1.6h   \\
      PMOCO-Aug  & \textbf{0.4712} & 0.00\%   & 4.9m  & 0.4409 & 0.63\%   & 28m   & 0.4956 & 1.82\%    & 1.6h   \\
      \midrule
      CNH-Aug    & 0.4704 & 0.17\%   & 8.5m  & 0.4409 & 0.63\%   & 28m   & 0.4996 & 1.03\%    & 1.6h   \\
      \textbf{POCCO-C-Aug}    & 0.4706 & 0.13\%   & 8.9m  & 0.4419 & 0.41\%   & 34m   & 0.5023 & 0.50\%    & 2h     \\
      \midrule
      WE-CA-Aug  & \textbf{0.4712} & 0.00\%   & 8.2m  & 0.4432 & 0.11\%   & 29m   & 0.5035 & 0.26\%    & 1.7h   \\
      \textbf{POCCO-W-Aug}    & \textbf{0.4712} & 0.00\%   & 8.9m  & \underline{0.4437} & 0.00\%   & 33m   & \underline{0.5048} & 0.00\%    & 2h     \\
\bottomrule
 \vspace{-4mm}
\end{tabular}}
\end{small}
\end{table}

\subsection{Experimental results}

\textbf{Comparison analysis.} The comparison results are presented in Table~\ref{tab:bitsp} and Table~\ref{tab:tritsp}. POCCO-W consistently achieves superior performance over WE-CA across all benchmark scenarios, establishing itself as the new SOTA results among neural MOCOP solvers. Similarly, POCCO-C outperforms CNH in every case. Both variants also surpass their augmentation-based counterparts, WE-CA-Aug and CNH-Aug, on Bi-TSP20 and Bi-CVRP100, highlighting POCCO’s enhanced ability to explore the solution space and approximate high-quality Pareto fronts. When further combined with instance augmentation, POCCO demonstrates additional performance gains. 
Please note that POCCO with instance augmentation yields lower HV values compared with its non-augmented counterpart on Bi-TSP20. This is because decomposition-based methods focus on optimizing individual subproblems rather than ensuring overall solution diversity. While augmentation improves solution quality for specific subproblems, it may reduce the number of non-dominated solutions, resulting in a smaller HV.
Compared with multi-model approaches that require training or fine-tuning separate models for each subproblem, POCCO delivers superior results while maintaining a single shared model. 
% It performs slightly below EMNH on Bi-KP100 but remains superior in most other cases.
Notably, POCCO achieves better results on Bi-TSP50 than WS-LKH, a setting where previous neural solvers have consistently failed. In terms of efficiency, POCCO significantly reduces computational time. For example, POCCO-W-Aug solves Bi-TSP100 in only 14 minutes, while WE-LKH requires about 6.0 hours, with POCCO delivering comparable solution quality.

\begin{figure}[!t]
  \centering
  \vspace{-0.8mm}
  \begin{subfigure}[b]{0.3\textwidth}
    \centering
    \includegraphics[width=\linewidth]{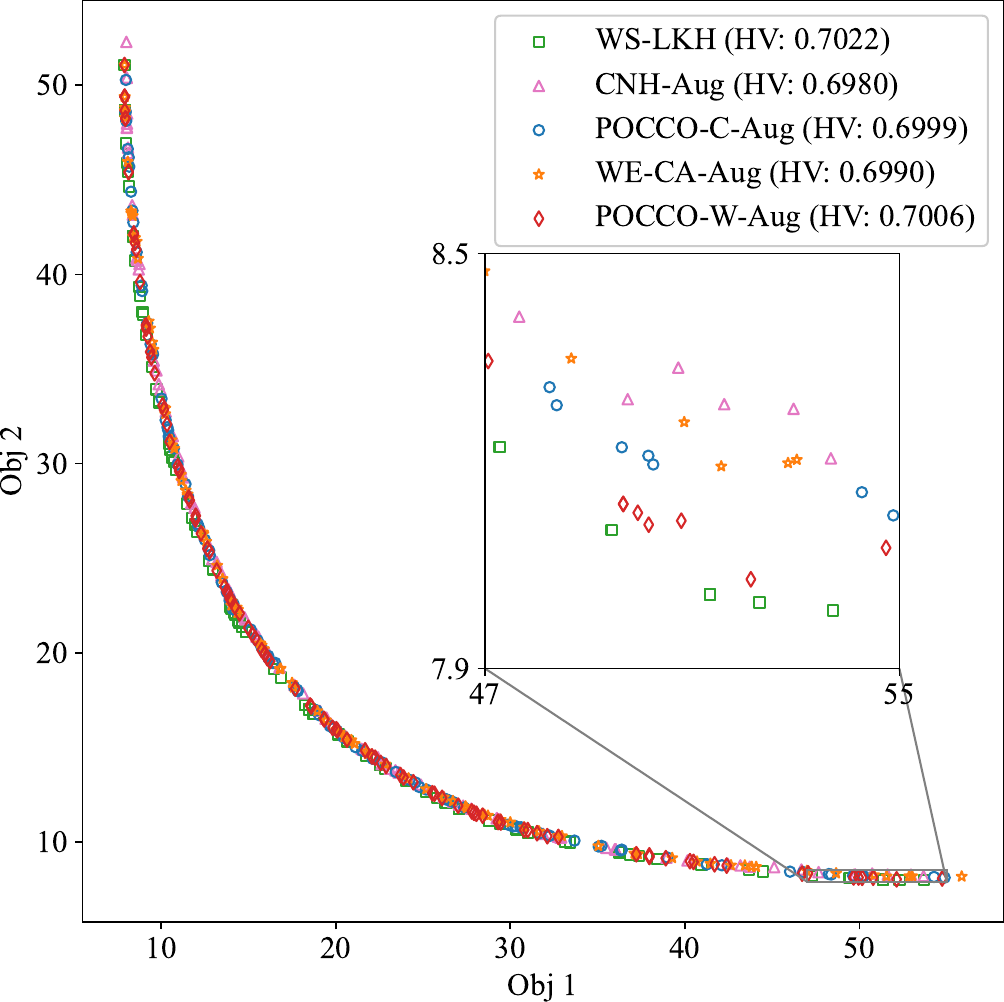}
    % \vspace{-3mm}
    \caption{KroAB100}
    \label{fig:pf100}
  \end{subfigure}%
  \hfill
  \begin{subfigure}[b]{0.3\textwidth}
    \centering
    \includegraphics[width=\linewidth]{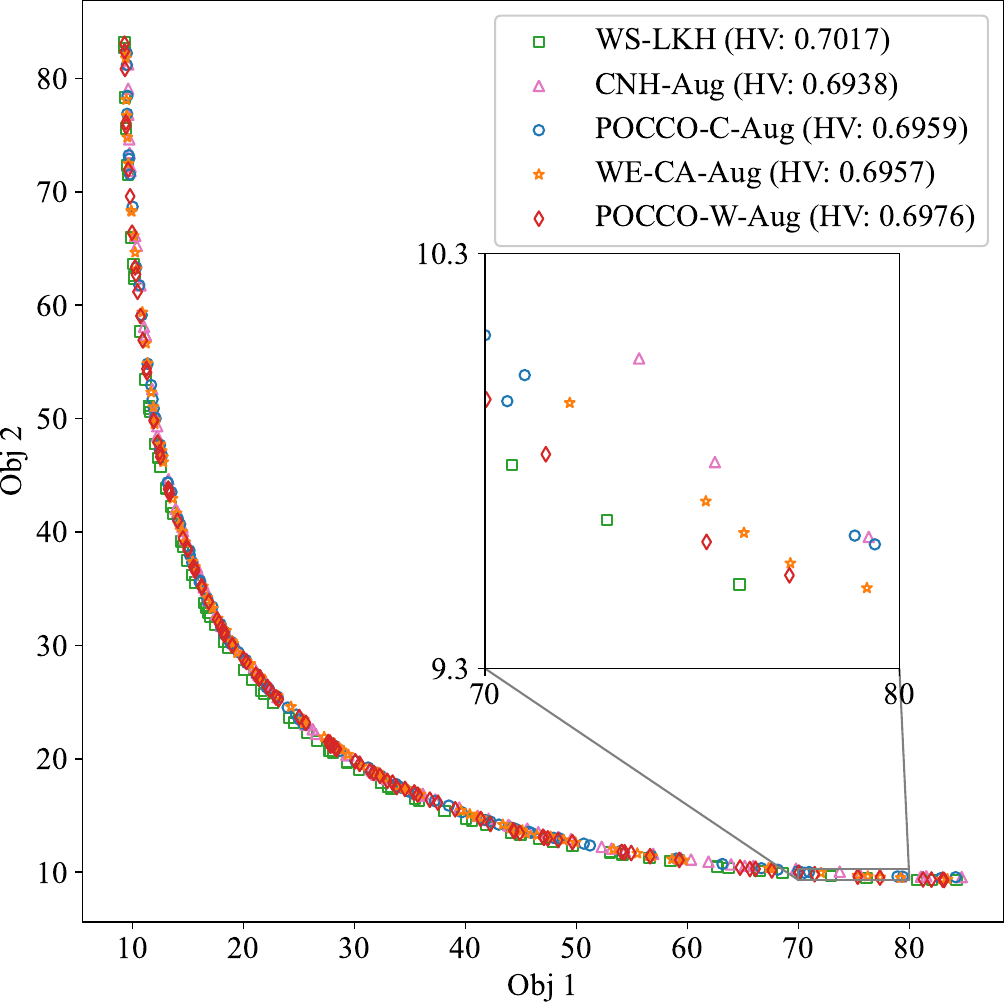}
    % \vspace{-3mm}
    \caption{KroAB150}
    \label{fig:pf150}
  \end{subfigure}%
  \hfill
  \begin{subfigure}[b]{0.3\textwidth}
    \centering
    \includegraphics[width=\linewidth]{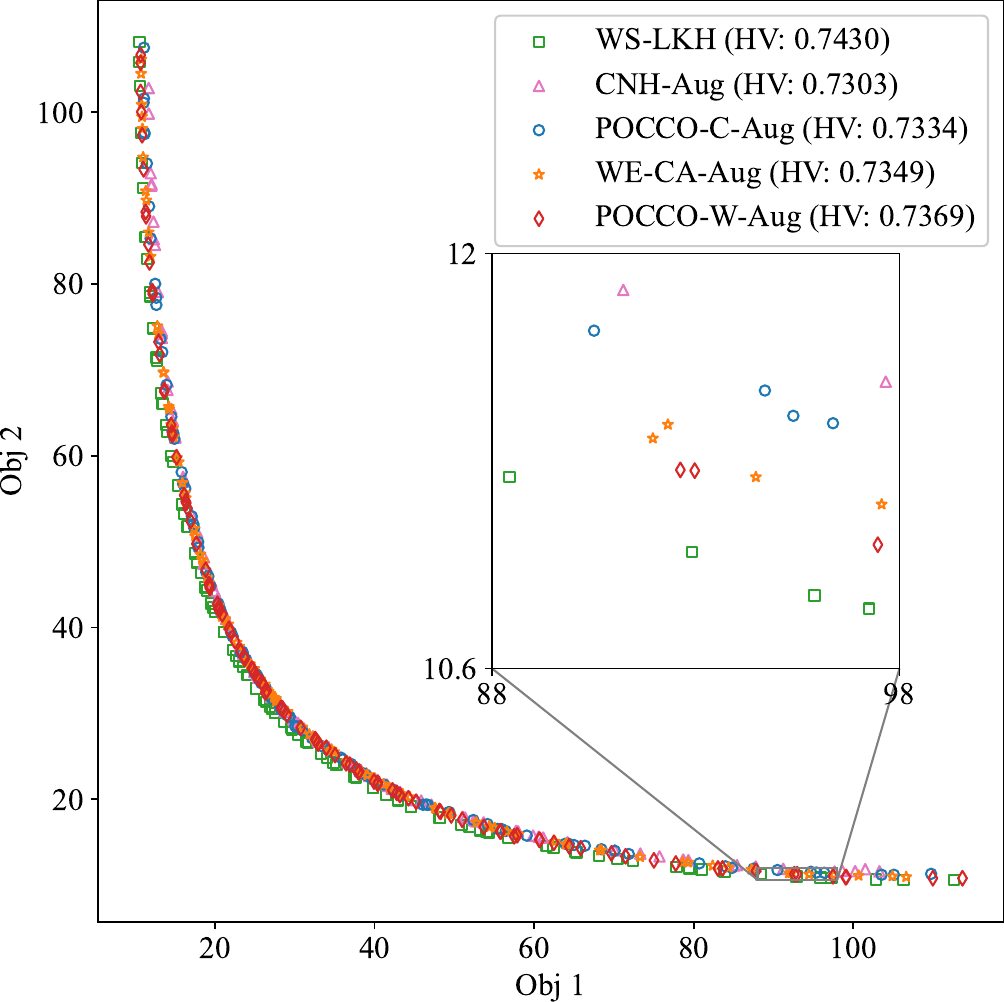}
    % \vspace{-3mm}
    \caption{KroAB200}
    \label{fig:pf200}
  \end{subfigure}
  % \vspace{-1mm}
  \caption{Pareto fronts of benchmark instances.}
  \label{fig:three_plots}
\end{figure}

\textbf{Out-of-distribution size generalization analysis.}
We evaluate the generalization ability of the models on out-of-distribution, benchmark instances KroAB100/150/200~\cite{lust2010two}. All neural methods are trained on Bi-TSP100, except for CNH, WE-CA, and POCCO, which are trained across varying sizes $n \in \{20, 21, \dots, 100\}$. The results, visualized in Fig.~\ref{fig:three_plots}, show that POCCO-W-Aug consistently achieves the best generalization performance compared with other neural baselines. POCCO-C-Aug also outperforms CNH-Aug across all evaluated scenarios. 
% The Pareto fronts in Fig.~\ref{fig:three_plots} clearly illustrate the superiority of POCCO-W and POCCO-C over WE-CA and CNH benchmark instances KroAB100/150/200. 
Full benchmark results and experiments on larger-scale instances Bi-TSP150/200 are provided in Appendix~\ref{appendix: benchmark} and Appendix~\ref{appendix:large_size}, respectively.

\begin{figure}[!t]
  \centering
  \begin{subfigure}[b]{0.33\textwidth}
    \centering
    \includegraphics[width=\linewidth]{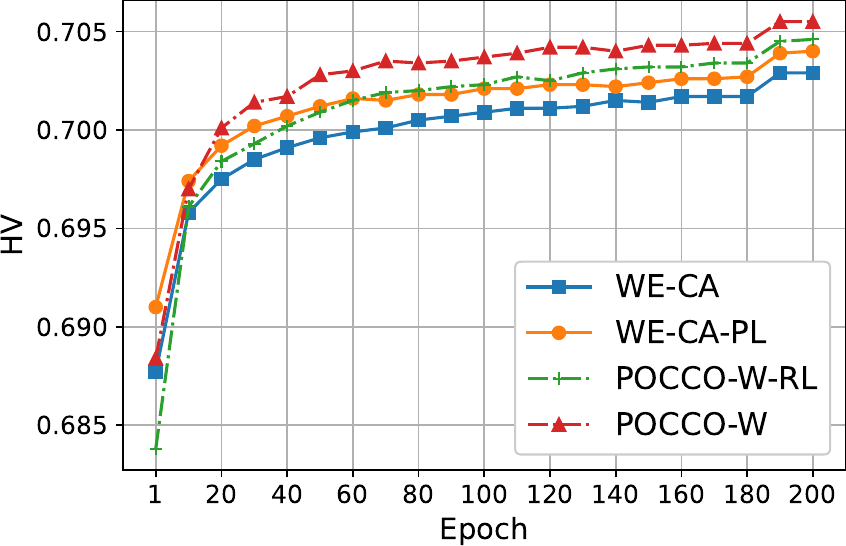}
    % \vspace{-3mm}
    \caption{Validation curves}
    \label{fig:val_cur}
  \end{subfigure}%
  \hfill
  \begin{subfigure}[b]{0.33\textwidth}
    \centering
    \includegraphics[width=\linewidth]{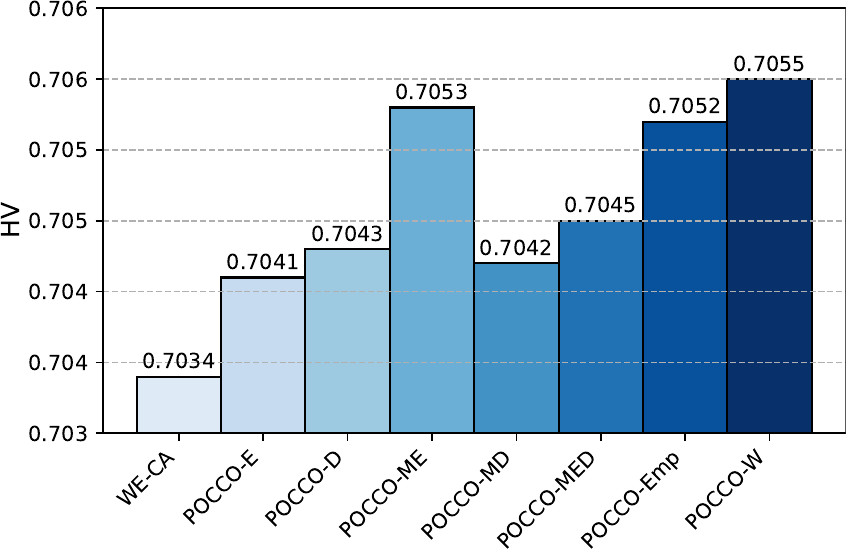}
    % \vspace{-3mm}
    \caption{Bi-TSP100}
    \label{fig:abla_100}
  \end{subfigure}%
  \hfill
  \begin{subfigure}[b]{0.33\textwidth}
    \centering
    \includegraphics[width=\linewidth]{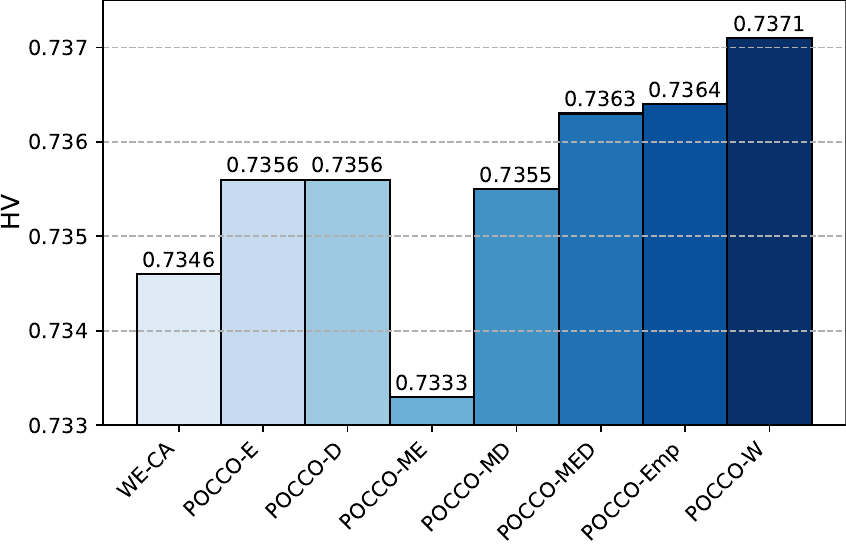}
    % \vspace{-3mm}
    \caption{Bi-TSP200}
    \label{fig:abla_200}
  \end{subfigure}
  % \vspace{-1mm}
  \caption{Ablation study:(a) validates the effectiveness of PL; (b) and (c) verify the effects of different CCO block variants.}
  \label{fig:ablation_study}
\end{figure}

%\textbf{Effectiveness of the PL.} We assess the training efficiency of the PL by comparing it to REINFORCE on the WE-CA and POCCO-W models using the Bi-TSP100 dataset. As shown by the validation curves in Fig. \ref{fig:val_cur}, PL achieves faster convergence despite identical network architectures. Notably, for WE-CA, training with PL for 100 epochs reaches performance comparable to 200 epochs of REINFORCE. Similar improvements are observed for POCCO-W. These results demonstrate that PL effectively accelerates the training process and achieves better performance with fewer training epochs.
\textbf{Effectiveness of PL.} We assess the training efficiency of PL by comparing it to REINFORCE on the WE-CA and POCCO-W models using the Bi-TSP100 dataset. As shown by the validation curves in Fig.~\ref{fig:val_cur}, PL achieves faster convergence despite identical network architectures. Notably, for WE-CA, training with PL for 100 epochs reaches performance comparable to 200 epochs of REINFORCE. Similar improvements are observed for POCCO-W. These results demonstrate that PL accelerates the training process and achieves better performance with fewer training epochs.

\textbf{Effectiveness of the CCO block.}
To evaluate the impact of the CCO block’s structure and placement, we compare POCCO-W with WE-CA and several POCCO variants: POCCO-E (CCO inserted in the encoder replacing the FF layer), POCCO-D (CCO replacing the final linear layer of MHA in the decoder, using MLP experts), POCCO-ME (replacing CCO with a standard MoE layer), POCCO-MD (replacing CCO with three MoD layers), POCCO-MED (using MoE in the encoder and MoD in place of CCO in the decoder), and POCCO-Emp (replacing the identity expert in CCO with an empty expert).
As shown in Fig. \ref{fig:abla_100}, all variants outperform WE-CA on the in-distribution Bi-TSP100, with POCCO-W, POCCO-ME, and POCCO-Emp achieving the most notable gains. On the out-of-distribution Bi-TSP200 in Fig. \ref{fig:abla_200}, only POCCO-W and POCCO-Emp maintain strong performance, while POCCO-ME performs worst, even underperforming WE-CA. These results highlight the importance of both the structure and placement of the CCO block for achieving strong generalization across in- and out-of-distribution settings.

% \begin{table}[htbp]
% \centering
% \caption{Performance of WE-CA variants on BiTSP Instances}
% \begin{small}
% \renewcommand\arraystretch{0.9}
% \resizebox{0.98\textwidth}{!}{ 
% \begin{tabular}{lccc ccc ccc}
% \toprule
% \multirow{2}{*}{Method} & \multicolumn{3}{c}{BiTSP20} & \multicolumn{3}{c}{BiTSP50} & \multicolumn{3}{c}{BiTSP100} \\
% \cmidrule(lr){2-4} \cmidrule(lr){5-7} \cmidrule(lr){8-10}
% & HV & Gap & Time & HV & Gap & Time & HV & Gap & Time \\
% \midrule
% WE-CA         & 0.6270 & 0.03\%  & 7s   & 0.6392 & 0.16\%  & 10s  & 0.7034 & 0.10\%  & 21s \\
% WE-CA-wo-ADA      & 0.6271 & 0.02\%  & 7s   & 0.6398 & 0.06\%  & 10s  & 0.7037 & 0.06\%  & 21s \\
% WE-CA-P   & 0.6272 & 0.00\%  & 7s   & 0.6402 & 0.00\%  & 12s  & 0.7041 & 0.00\%  & 23s \\
% \bottomrule
% \end{tabular}}
% \end{small}
% \end{table}

\textbf{Hyperparameter study.}
We conduct experiments to examine the impact of different key hyperparameters on POCCO’s performance. As detailed in Appendix~\ref{appendix: hyperparam}, the number of CCO block layers, the number of experts, the $\operatorname{Top}k$ value, and the temperature parameter $\beta$ all influence model effectiveness. For the problems studied, the desirable settings, as identified based on empirical results, are: one CCO block layer, four FF experts and one ID expert, $\operatorname{Top}k=2$, $\beta=3.5$ for bi-objective tasks, and $\beta=4.5$ for tri-objective tasks.

\section{Conclusion}
\label{sec:conclu}
This paper presents POCCO, a plug-and-play framework tailored for MOCOPs, which adaptively routes subproblems through different model structures and leverages PL for more effective training. 
% We first introduce a CCO block that enables the model to learn a diverse policy ensemble, with each policy specialized for a subset of subproblems, thereby improving overall solution quality. We then develop a preference-driven learning strategy based on pairwise comparisons, which promotes efficient exploration in the vast solution space and enhances training stability. 
POCCO is integrated into two SOTA neural solvers, and extensive experiments demonstrate its effectiveness. Ablation studies further highlight the necessity of both CCO block and PL, and reveal the critical impact of the design and placement of the CCO block. 
We acknowledge certain limitations, such as the limited capability to address real-world MOCOPs with complex constraints or large problem sizes. Addressing these challenges may require constraint-handling mechanisms \cite{bi2024learning} or divide-and-conquer \cite{li2021learning} strategies, which we leave for future work.
% To better , 
% we plan to incorporate constraint-handling mechanisms. Additionally, to scale POCCO to solve large-scale instances, efficient techniques such as divide-and-conquer may be needed. 
% we aim to integrate generative modeling techniques to avoid expensive autoregressive decoding.

\begin{ack}
We thank the anonymous reviewers and (S)ACs of NeurIPS 2025 for their constructive comments and dedicated service to the community.
\end{ack}

% \clearpage
{
\small
\bibliographystyle{plain}
\bibliography{arxiv}
}

%%%%%%%%%%%%%%%%%%%%%%%%%%%%%%%%%%%%%%%%%%%%%%%%%%%%%%%%%%%%
% \newpage
% \appendix

% \vbox{
% \hrule height 4pt
% \vskip 0.25in
% \vskip -\parskip%
% \centering
% {\LARGE\bf Appendix} \\
% % \vskip 0.1in
% % {\LARGE\bf Optimization with Conditional Computation (Appendix)}
% \vskip 0.29in
% \vskip -\parskip
% \hrule height 1pt
% }

% \section{XXX}
% XXX.
%%%%%%%%%%%%%%%%%%%%%%%%%%%%%%%%%%%%%%%%%%%%%%%%%%%%%%%%%%%%
\newpage
\section*{NeurIPS Paper Checklist}

\begin{enumerate}

\item {\bf Claims}
    \item[] Question: Do the main claims made in the abstract and introduction accurately reflect the paper's contributions and scope?
    \item[] Answer: \answerYes{} % Replace by \answerYes{}, \answerNo{}, or \answerNA{}.
    \item[] Justification: Main contributions and scope are accurately reflect in Abstract and Section \ref{sec:intro}.

\item {\bf Limitations}
    \item[] Question: Does the paper discuss the limitations of the work performed by the authors?
    \item[] Answer: \answerYes{} % Replace by \answerYes{}, \answerNo{}, or \answerNA{}.
    \item[] Justification: The limitations are discussed in Section \ref{sec:conclu}.

\item {\bf Theory assumptions and proofs}
    \item[] Question: For each theoretical result, does the paper provide the full set of assumptions and a complete (and correct) proof?
    \item[] Answer: \answerNA{} % Replace by \answerYes{}, \answerNo{}, or \answerNA{}.
    \item[] Justification: The paper does not include new theoretical results.

    \item {\bf Experimental result reproducibility}
    \item[] Question: Does the paper fully disclose all the information needed to reproduce the main experimental results of the paper to the extent that it affects the main claims and/or conclusions of the paper (regardless of whether the code and data are provided or not)?
    \item[] Answer: \answerYes{} % Replace by \answerYes{}, \answerNo{}, or \answerNA{}.
    \item[] Justification: The paper fully discloses all the information needed to reproduce the main experimental results in Section \ref{sec:experiments}.

\item {\bf Open access to data and code}
    \item[] Question: Does the paper provide open access to the data and code, with sufficient instructions to faithfully reproduce the main experimental results, as described in supplemental material?
    \item[] Answer: \answerNo{} % Replace by \answerYes{}, \answerNo{}, or \answerNA{}.
    \item[] Justification: We promise that the source code and data will be publicly released with the MIT License upon publication.

\item {\bf Experimental setting/details}
    \item[] Question: Does the paper specify all the training and test details (e.g., data splits, hyperparameters, how they were chosen, type of optimizer, etc.) necessary to understand the results?
    \item[] Answer: \answerYes{} % Replace by \answerYes{}, \answerNo{}, or \answerNA{}.
    \item[] Justification: The detailed experimental setups are presented in Section \ref{sec:experiments}.

\item {\bf Experiment statistical significance}
    \item[] Question: Does the paper report error bars suitably and correctly defined or other appropriate information about the statistical significance of the experiments?
    \item[] Answer: \answerNo{} % Replace by \answerYes{}, \answerNo{}, or \answerNA{}.
    \item[] Justification: Error bars are not reported because it would be too computationally expensive.

\item {\bf Experiments compute resources}
    \item[] Question: For each experiment, does the paper provide sufficient information on the computer resources (type of compute workers, memory, time of execution) needed to reproduce the experiments?
    \item[] Answer: \answerYes{} % Replace by \answerYes{}, \answerNo{}, or \answerNA{}.
    \item[] Justification: Experiments compute resources are indicated in Section \ref{sec:experiments}.
    
\item {\bf Code of ethics}
    \item[] Question: Does the research conducted in the paper conform, in every respect, with the NeurIPS Code of Ethics \url{https://neurips.cc/public/EthicsGuidelines}?
    \item[] Answer: \answerYes{} % Replace by \answerYes{}, \answerNo{}, or \answerNA{}.
    \item[] Justification: This paper respects the NeurIPS Code of Ethics.

\item {\bf Broader impacts}
    \item[] Question: Does the paper discuss both potential positive societal impacts and negative societal impacts of the work performed?
    \item[] Answer: \answerYes{} % Replace by \answerYes{}, \answerNo{}, or \answerNA{}.
    \item[] Justification: The potential positive societal impacts and negative societal impacts of this work are discussed in Appendix \ref{appendix:impact}.
    
\item {\bf Safeguards}
    \item[] Question: Does the paper describe safeguards that have been put in place for responsible release of data or models that have a high risk for misuse (e.g., pretrained language models, image generators, or scraped datasets)?
    \item[] Answer: \answerNA{} % Replace by \answerYes{}, \answerNo{}, or \answerNA{}.
    \item[] Justification: The paper poses no such risks.

\item {\bf Licenses for existing assets}
    \item[] Question: Are the creators or original owners of assets (e.g., code, data, models), used in the paper, properly credited and are the license and terms of use explicitly mentioned and properly respected?
    \item[] Answer: \answerYes{} % Replace by \answerYes{}, \answerNo{}, or \answerNA{}.
    \item[] Justification: Licenses for existing assets are detailed in Appendix \ref{appendix:assets}.

\item {\bf New assets}
    \item[] Question: Are new assets introduced in the paper well documented and is the documentation provided alongside the assets?
    \item[] Answer: \answerNA{} % Replace by \answerYes{}, \answerNo{}, or \answerNA{}.
    \item[] Justification: The paper does not release new assets.

\item {\bf Crowdsourcing and research with human subjects}
    \item[] Question: For crowdsourcing experiments and research with human subjects, does the paper include the full text of instructions given to participants and screenshots, if applicable, as well as details about compensation (if any)? 
    \item[] Answer: \answerNA{} % Replace by \answerYes{}, \answerNo{}, or \answerNA{}.
    \item[] Justification: The paper does not involve crowdsourcing nor research with human subjects.

\item {\bf Institutional review board (IRB) approvals or equivalent for research with human subjects}
    \item[] Question: Does the paper describe potential risks incurred by study participants, whether such risks were disclosed to the subjects, and whether Institutional Review Board (IRB) approvals (or an equivalent approval/review based on the requirements of your country or institution) were obtained?
    \item[] Answer: \answerNA{} % Replace by \answerYes{}, \answerNo{}, or \answerNA{}.
    \item[] Justification: The paper does not involve crowdsourcing nor research with human subjects.

\item {\bf Declaration of LLM usage}
    \item[] Question: Does the paper describe the usage of LLMs if it is an important, original, or non-standard component of the core methods in this research? Note that if the LLM is used only for writing, editing, or formatting purposes and does not impact the core methodology, scientific rigorousness, or originality of the research, declaration is not required.
    %this research? 
    \item[] Answer: \answerNA{} % Replace by \answerYes{}, \answerNo{}, or \answerNA{}.
    \item[] Justification: The core method development in this research does not involve LLMs as any important, original, or non-standard components.

\end{enumerate}

%%%%%%%%%%%%%%%%%%%%%%%%%%%%%%%%%%%%%%%%%%%%%%%%%%%%%%%%%%%%
\newpage
% \documentclass{article}

% % if you need to pass options to natbib, use, e.g.:
% %     \PassOptionsToPackage{numbers, compress}{natbib}
% % before loading neurips_2025

% % ready for submission
% \usepackage{neurips_2025}

% % to compile a preprint version, e.g., for submission to arXiv, add add the
% % [preprint] option:
% %     \usepackage[preprint]{neurips_2025}

% % to compile a camera-ready version, add the [final] option, e.g.:
% %     \usepackage[final]{neurips_2025}

% % to avoid loading the natbib package, add option nonatbib:
% %    \usepackage[nonatbib]{neurips_2025}

% \usepackage[utf8]{inputenc} % allow utf-8 input
% \usepackage[T1]{fontenc}    % use 8-bit T1 fonts
% \usepackage{hyperref}       % hyperlinks
% \usepackage{url}            % simple URL typesetting
% \usepackage{booktabs}       % professional-quality tables
% \usepackage{multirow}

% \usepackage{graphicx}

% \usepackage{amsfonts}       % blackboard math symbols
% \usepackage{amsmath}
% \usepackage{nicefrac}       % compact symbols for 1/2, etc.
% \usepackage{microtype}      % microtypography
% \usepackage{xcolor}         % colors

% % \title{Appendix}

% \begin{document}
% \maketitle

\appendix

\vbox{
\hrule height 4pt
\vskip 0.25in
\vskip -\parskip%
\centering
{\LARGE\bf Appendix}
\vskip 0.29in
\vskip -\parskip
\hrule height 1pt
}

\section{Details of MOCOP}
Here we elaborate on the problem definitions for the three typical MOCOPs, i.e., MOTSP, MOCVRP, and MOKP.

\noindent\textbf{MOTSP.} An MOTSP instance involves multiple cost matrices, aiming to identify a set of tours, i.e., node sequences, that are Pareto optimal. For instance, a $\kappa$-objective TSP instance $\mathcal{G}$ with $n+1$ nodes is featured by 
%$m\,n \times n$ 
cost matrices $C^{i} = (c_{j,k}^{i}),$ with $ i \in \left\{ 1, \cdots, \kappa \right\}$ and $j,k \in \left\{0, \cdots, n\right\}$. The $\kappa$ objectives are defined as follows,
\begin{equation}
\begin{split}
    &\min_{\pi\in \mathcal{X}} F(\pi) = \min (f_{1}(\pi), f_{2}(\pi), \cdots, f_{\kappa}(\pi)), \\
    &{\rm with}\; f_{i}(\pi) = c_{\pi_{n}\pi_{0}}^{i} + \sum_{j=0}^{n-1}c_{\pi_{j}\pi_{j+1}}^{i},
\end{split}   
\end{equation}
where $\pi=\left(\pi_{0}, \pi_{2}, \cdots, \pi_{n}\right)$ with $\pi_{j}\!\in \! \left\{0, \cdots, n\right\}$. $\mathcal{X}$ represents all feasible solutions (i.e., tours), ensuring each node is visited exactly once. This paper considers the Euclidean MOTSP following \cite{florios2014generation,likaiwen2020deep,lin2022pareto}. Each node $j$ has a 2$\kappa$-dim feature vector $o_j=[loc_{j}^{1}, loc_{j}^{2}, \cdots, loc_{j}^{\kappa}]$, where $loc_{j}^{i} \in \mathbb{R}^{2}$ is the coordinate for the $i$-th objective. The objective $f_{i}(\pi)$ is defined as $f_{i}(\pi) = \|loc_{\pi_{n}}^{i}-loc_{\pi_{0}}^{i}\|_{2}+ \sum_{j=0}^{n-1}\|loc_{\pi_{j}}^{i}-loc_{\pi_{j+1}}^{i}\|_{2}$.

\noindent\textbf{MOCVRP.} An MOCVRP instance consists of $n$ customer nodes and one depot node. Each node has a 3-dim feature vector $o_j=[loc_{j}, \delta_{j}]$, where $loc_{j}$ and $\delta_{j}$, for $j\in \left\{0, \cdots, n\right\}$, correspond to the coordinates and demand of node $j$. Notably, for the depot node, $o_0=[loc_{0}, \delta_{0}]$, the demand $\delta_{0}$ is set to 0. Vehicles with a capacity of $\mathcal{Q}$ ($\mathcal{Q}>\delta_{i}$) are employed to serve all customers in multiple routes, with each route commencing and concluding at the depot. The problem must satisfy the following constraints: 1) Each customer is visited exactly once, and 2) The total demand of customers in each route must not exceed the vehicle's capacity.
In our study, we focus on the bi-objective CVRP, aligning with prior research \cite{lacomme2006genetic,lin2022pareto}. Specifically, we aim to minimize two objectives: the total length of all routes and the length of the longest route (i.e., the makespan).

\noindent\textbf{MOKP.}
An MOKP instance consists of $n+1$ items, where each item $j$ is characterized by a 2-dim feature vector $o_j = [w_j, p_j]$, representing its weight $w_j$ and profit vector $p_j$, for $j \in \{0, \cdots, n\}$. The profit vector $p_j \in \mathbb{R}^\kappa$ corresponds to $\kappa$ distinct objective values associated with item $j$. A knapsack with a capacity of $\mathcal{C}$ is provided, where $\mathcal{C} > w_j$ for each item. The goal is to select a subset of items to place into the knapsack while satisfying the following constraint: the total weight of selected items must not exceed the knapsack capacity $\mathcal{C}$.
In our study, we focus on the bi-objective knapsack problem, consistent with previous research \cite{lust2010two,zhou2012multiobjective}. Specifically, we aim to simultaneously maximize two objectives: the sum of the first and second profit components across the selected items.

\section{Theoretical Analysis}

We derive the loss function and gradient for our PL objective and contrast it with the REINFORCE gradient. Our theoretical (and empirical) analyses demonstrate that PL yields significantly lower gradient variance, which contributes to faster and more stable convergence. 

\subsection{Formulation of Loss Function}

Different from the RL, our PL method exploits the preference relations between generated solutions according to their objective. The explicit preference $f^* (\pi|\mathcal{G}, \lambda)$ is defined as the negation of the scalarized objective. We contruct a preference pair, denoted as $(\pi^w, \pi^\ell)$, along with a binary preference label $y$, where $y = 1$ if $\pi^{w} \lessdot \pi^{l}$ (i.e., $f^* (\pi^w|\mathcal{G}, \lambda) > f^* (\pi^\ell|\mathcal{G}, \lambda)$), and $y = 0$ otherwise.

For the policy $p_\theta(\pi|\mathcal{G}, \lambda)$ used to construct solution $\pi$, its implicit preference is defined as the average log-likelihood: $f_\theta (\pi\mid \mathcal{G}, \lambda)= \frac{1}{|\pi|} \log p_{\theta} (\pi\,|\,\mathcal{G}, \lambda).$ For a preference pair $(\pi^w, \pi^\ell)$, the preference distribution is modeled using the Bradley-Terry (BT) ranking objective and implicit preferences:

$$
g_{\theta} \bigl (\pi^{w} \lessdot \pi^{\ell}\mid\mathcal{G},\lambda \bigr)=
\sigma \bigl( \beta\,[\,f_{\theta}(\pi^{w}\mid\mathcal{G},\lambda) - f_{\theta}(\pi^{\ell}\mid\mathcal{G},\lambda)\,]\bigr),
$$

where $\sigma(\cdot)$ is the sigmoid function, and $\beta>0$ is a fixed temperature that controls the sharpness with which the model distinguishes between unequal rewards. By maximizing the log-likelihood of $g_{\theta} \bigl (\pi^{w} \lessdot \pi^{\ell}\mid\mathcal{G},\lambda \bigr)$, the model is encouraged to assign higher probabilities to preferred solutions $\pi^w$ compared with less preferred solutions $y^\ell$. We can derive the PL loss function:

$$
\mathcal{L}_ {PL} (\theta|p_\theta, \mathcal{G},\lambda, \pi^w, \pi^l) = - y\,\log\, \sigma \bigl( \beta [\frac{\log\, p_\theta(\pi^w|\mathcal{G},\lambda)}{|\pi^w|}-\frac{\log\, p_\theta(\pi^l|\mathcal{G},\lambda)}{|\pi^l|}\bigr]),
$$

PL directly distinguishes optimization signals based on exact preferences, and the strength of the optimization signal correlates with the difference in log-likelihood, requiring the model to maximize the probability gap between $\pi^w$ and $\pi^\ell$. 

\subsection{Gradient Analysis}

Let $z$ denote the argument of the sigmoid function:
$$
z=\beta [\frac{\log\, p_\theta(\pi^w|\mathcal{G},\lambda)}{|\pi^w|}-\frac{\log\, p_\theta(\pi^l|\mathcal{G},\lambda)}{|\pi^l|}\bigr].
$$

The gradient of $\mathcal{L}_ {PL}$ with respect to $\theta$ is:
$$
\nabla_\theta \mathcal{L}_ {PL} = \frac{\partial \mathcal{L}_ {PL}}{\partial z} \cdot \nabla_\theta z.
$$

Derivative of $-y \log \sigma(z)$ becomes:
$$
\frac{\partial \mathcal{L}_ {PL}}{\partial z} = - y(1 - \sigma(z)).
$$

Gradient of $z$ with respect to $\theta$:
$$
\nabla_\theta z = \beta [\frac{1}{|\pi^w|} \nabla_\theta \log p_\theta(\pi^w|\mathcal{G},\lambda) - \frac{1}{|\pi^\ell|} \nabla_\theta \log p_\theta(\pi^\ell | \mathcal{G},\lambda)].
$$

Combining these, the total gradient becomes:
$$
\nabla_\theta \mathcal{L}_ {PL}= - y \beta(1 - \sigma(z))[\frac{1}{|\pi^w|} \nabla_\theta \log p_\theta(\pi^w|\mathcal{G},\lambda) - \frac{1}{|\pi^\ell|} \nabla_\theta \log p_\theta(\pi^\ell | \mathcal{G},\lambda)].
$$

The gradient increases the likelihood of $\pi^w$ and decreases the likelihood of $\pi^\ell$. For comparison, we analyze the loss function in the REINFORCE algorithm here:
$$
\mathcal{L}_ {RL} (\pi|\mathcal{G}, \lambda)=-(\mathcal{R}(\pi)-b) \text{log} p_{\theta} (\pi|\mathcal{G},\lambda),
$$

where $b$ is a baseline to distinguish positive or negative optimization signals for each solution $\pi$. The gradient of the REINFORCE algorithm is:

$$
\nabla_\theta \mathcal{L}_ {RL} = - (\mathcal{R}(\pi)-b) \nabla_\theta \text{log} p_\theta (\pi|\mathcal{G},\lambda).
$$

REINFORCE relies on \emph{absolute} returns $\mathcal{R}(\pi)$, whose large variance propagates directly to the gradient. In contrast, $\mathcal{L}_{\mathrm{PL}}$ uses \emph{relative} returns inside a sigmoid, yielding the difference of two normalized log-likelihoods. This pairwise, centered signal reduces gradient variance and produces smoother, more stable updates, leading to faster and more reliable convergence. This theoretical insight is further supported by our empirical analyses.  We provide concrete evidence of this in Table~\ref{tab:graident}, which reports the average gradient variance in the first five batches of training POCCO-W under both RL and PL frameworks. Across three MOTSP settings, PL consistently exhibits two to four orders of magnitude lower variance, leading to faster convergence and more stable optimization dynamics.

\begin{table}[htbp]
    \centering
    \caption{Gradient variance in REINFORCE vs. preference learning.}
    \label{tab:graident}
    \begin{tabular}{c c c c}
        \toprule
        \textbf{Problem Size} & \textbf{Batch} & \textbf{RL Variance} & \textbf{PL Variance} \\
        \midrule
        MOTSP20  & 1 & 0.054648 & 0.000314 \\
                 & 2 & 0.039951 & 0.000252 \\
                 & 3 & 0.019038 & 0.000191 \\
                 & 4 & 0.009093 & 0.000115 \\
                 & 5 & 0.010742 & 0.000104 \\
        \midrule
        MOTSP50  & 1 & 0.474784 & 0.000142 \\
                 & 2 & 0.220530 & 0.000077 \\
                 & 3 & 0.140275 & 0.000048 \\
                 & 4 & 0.092114 & 0.000040 \\
                 & 5 & 0.065286 & 0.000031 \\
        \midrule
        MOTSP100 & 1 & 10.124832 & 0.000059 \\
                 & 2 & 6.355499  & 0.000039 \\
                 & 3 & 3.461700  & 0.000032 \\
                 & 4 & 1.755804  & 0.000021 \\
                 & 5 & 1.111246  & 0.000014 \\
        \bottomrule
    \end{tabular}
\end{table}

\section{MOCOP Solvers with POCCO}
\label{app:b}

Existing neural MOCOP solvers are based on a transformer-based architecture that consists of an encoder and a decoder. The encoder is used to generate embeddings for all nodes based on the instance $\mathcal{G}$ and the weight vector $\lambda$. The decoder is used to decode a sequence of actions based on these embeddings in an iterative fashion. To demonstrate the versatility of our POCCO, we integrate it into two SOTA neural methods, WE-CA~\cite{chen2025rethinking} and CNH~\cite{fan2024conditional}, yielding POCCO-W and POCCO-C.

\subsection{POCCO-W}
Given an instance \(\mathcal{G}\) comprising \(n+1\) nodes with \(Z\)-dimensional features \(\{o_i\}_{i=0}^n\subset\mathbb R^Z\) and a weight vector $\lambda$,  first obtains initial embeddings by applying separate linear projections to the node features and the weight vector:
\begin{equation}
h_i^{0} = W^o\,o_i + b^o,\quad \forall\,i\in\{0,\dots,n\},\quad h_\lambda^{0} = W^\lambda\,\lambda + b^\lambda,
\label{ap_eq:linear}
\end{equation}
where \(W^o\in\mathbb R^{d\times Z}\), \(W^\lambda \in\mathbb R^{d\times \kappa}\), and \(b^o, b^\lambda \in\mathbb R^d\) are learnable parameters, with  embedding dimension \(d=128\).
POCCO-W integrates the weight embedding into each node embedding in a feature-wise fashion within the encoder. To ensure harmonious interaction, the weight and node embeddings are updated simultaneously. The encoder itself comprises $L=6$ transformer layers,  each layer applying a conditional attention sublayer, followed by a residual Add \& Norm (skip connection with instance normalization), a fully connected feed-forward sublayer, and a second Add \& Norm.

Specifically, the conditional attention sublayer first conditions node embeddings on the weight embedding via a feature‐wise affine transform:
\begin{align}
\gamma &= W^\gamma\,h_\lambda^{l-1}, 
\quad
\beta  = W^\beta\,h_\lambda^{l-1}, 
\quad
h_i' = \gamma \circ h_i^{l-1} + \beta,
\quad \forall\,i\in\{0,\dots,n\},
\label{ap_eq:feature_wise}
\end{align}
where $W^\gamma$ and $W^\beta$ are trainable matrices; $\circ$ denotes element‐wise multiplication. Then, the weight and node embeddings are updated via the  Multi‐Head Attention (MHA) mechanism with 8 heads and an Add \& Norm, as follows:
\begin{align}
\hat h_\lambda
&= \mathrm{IN}\bigl(h_\lambda^{l-1}
  + \mathrm{MHA}\bigl(h_\lambda^{l-1},\,\{h_\lambda^{l-1},h_0',\dots,h_n'\}\bigr)\bigr),
\label{ap_eq:W-lamda}\\
\hat h_i
&= \mathrm{IN}\bigl(h_i^{l-1}
  + \mathrm{MHA}\bigl(h_i',\,\{h_\lambda^{l-1},h_0',\dots,h_n'^\}\bigr)\bigr),
\quad \forall\,i \in\{0,\dots,n\}.
\label{ap_eq:W-node}
\end{align}
Afterwards, a fully connected feed-forward sublayer and another Add \& Norm are employed to yield the weight embedding $h_\lambda^{l}$ and the node embeddings 
$\{h_i^{l}\}_{i=0}^n$, as follows:
\begin{align}
h_i^{l} &= \mathrm{IN}\bigl(\hat h_i + \mathrm{FF}(\hat h_i)\bigr),
\label{ap_eq:w-ff} \\
h_\lambda^{l} &= \mathrm{IN}\bigl(\hat h_\lambda + \mathrm{FF}(\hat h_\lambda)\bigr),
\end{align}

Given the eventual node embeddings $\{h_i\}_{i=0}^n$ and weight embedding $h_\lambda$ output by the encoder, the decoder autoregressively computes the probability of node selection over $T$ steps. At decoding step \(t \in \{1, \dots, T\}\), the advanced context vector \(h_c\) is produced by an MHA layer with 8 heads based on the context embedding $v_c$ and eventual node embeddings as follows:

\begin{equation}
h_c \;=\; \mathrm{MHA}\Bigl(v_c,\;\{h_\lambda,\,h_0,\dots,h_n\}\Bigr),
\label{ap_eq:decoder_mha}
\end{equation}

\noindent\textbf{Context embedding.} For MOTSP, the context embedding $v_c$ is obtained by concatenating the embeddings of the first and last visited nodes, and all previously visited nodes are masked when computing selection probabilities. In MOCVRP, $v_c$ consists of the embedding of the last visited node together with the remaining vehicle capacity, while nodes that have already been visited or whose demand exceeds the remaining capacity are masked. In MOKP, the context embedding combines the graph embedding $\bar{h} = \frac{1}{n+1}\sum_{i=0}^n h_i$
with the remaining knapsack capacity, masking both items that are already selected and those whose weight exceeds the remaining capacity when computing selection probabilities.

The context vector $h_c$ is fed through the CCO block to produce a glimpse $g_c$ based on Eq.~\ref{eq:cco} (i.e., $g_c=\text{CCO}(h_c)$). This glimpse $g_c$  is then used in a compatibility layer to compute unormalized compatibility scores $\alpha$ as follows:
\begin{equation}
\alpha_i =
\begin{cases}
-\infty, & \text{if node }i\text{ is masked},\\
C\cdot\tanh\!\bigl(\tfrac{g_c^\top (W^K h_i)}{\sqrt{d}}\bigr), & \text{otherwise},
\end{cases}
\label{ap_eq:decoder_com}
\end{equation}
where \(C\) is set to 50 and $W^K$ is a learnable weight matrix. Finally, the node‐selection probability for the scalarized subproblem is given by:
\begin{equation}
    P_\theta(\pi_t \mid \pi_{1:t-1}, s) \;=\; \operatorname{Softmax}(\alpha).
\label{ap_eq:softmax}
\end{equation}

\subsection{POCCO-C}
POCCO-C mirrors the overall design of POCCO-W but (i) replaces every conditional-attention sublayer in the encoder with a dual-attention sublayer and (ii) enriches the context vector $h_c$ via a problem-size embedding (PSE) layer.   
Specifically, each of the $L$ encoder layers in POCCO-C consists of a dual-attention sublayer, followed by a residual Add \& Norm, a fully connected feed-forward sublayer, and a second Add \& Norm. Concretely, at layer $l$ the dual-attention and first Add \& Norm operate as:
\begin{align}
\hat h_\lambda
&= \mathrm{IN}\bigl(h_\lambda^{l-1}
  + \mathrm{MHA}\bigl(h_\lambda^{l-1},\,\{h_\lambda^{l-1},h_0^{l-1},\dots,h_n^{l-1}\}\bigr)\bigr),
\label{ap_eq:C_lamda}\\
\hat h_i'
&= \mathrm{MHA}\bigl(h_i^{l-1},\,\{h_\lambda^{l-1},h_0^{l-1},\dots,h_n^{l-1}\}\bigr) + \mathrm{MHA}\bigl(h_i^{l-1},\,\{h_\lambda^{l-1}\}\bigr),
\quad \forall\,i \in\{0,\dots,n\}.\\
\hat h_i
&= \mathrm{IN}\bigl(h_i^{l-1}
  + \hat h_i'\bigr),
\quad \forall\,i \in\{0,\dots,n\}.
\label{ap_eq:C_node}
\end{align}

Besides, POCCO-C employs the sinusoidal encoding based on sine and cosine functions to yield the PSE as follows,
\begin{equation}
\begin{split}
  &  \text{PSE}(\xi, 2i) = sin(\xi/10000^{2i/d}), \\
  &  \text{PSE}(\xi, 2i+1) = cos(\xi/10000^{2i/d}),
\end{split}
\end{equation}
where $\xi (\xi= n+1)$ and $i$ ($i\in \left\{0, \cdots, 63\right\}$) mean the problem size and dimension. The resulting d-dimensional PSEs are then processed using two linear layers with trainable matrices $W_{\xi1} \in \mathbb{R}^{d\times 2d}$ and $W_{\xi2} \in \mathbb{R}^{2d\times d}$. POCCO-C injects the size information by adding the results from PSE to $\{h_{i}\}_{i=0}^{n}$ from the encoder, such that,
\begin{equation} \label{eq:pse_linear}
    {h_{i}^{\xi}} = h_{i} + h_{\xi}, \; \text{with} \; h_{\xi} = (\text{PSE}(\xi, \cdot)W_{\xi1})W_{\xi2}.
\end{equation}

Then, POCCO-C produces the advanced context vector $h_c$ based on the context embedding $v_c$ and the size-injected node embeddings $\{h_i^\xi\}_{i=0}^n$ through a MHA layer with 8 heads as follows,
\begin{equation}
h_c \;=\; \mathrm{MHA}\Bigl(v_c,\;\{h_0^\xi,\dots,h_n^\xi\}\Bigr).
\label{ap_eq:decoder_mha_pocco}
\end{equation}

\subsection{Gating Mechanism}
% We employ subproblem-level gating, routing each context vector $h_c$ independently to a subset of experts. 
% Let \(d\) be the hidden dimension and \(W_G\in\mathbb R^{d\times (m+1)}\) the trainable gating weights in POCCO. Given a batch of context vectors $X=\{h_c^b\}_{b=1}^B, X\in\mathbb R^{B\times d}$,
% where $B$ is the batch size, the gating network computes the score matrix $H \;=\; X\,W_G \;\in\;\mathbb R^{B\times (m+1)}$.
% Subproblem \(b\) (with context vetor \(h_c^b\)) is routed to expert \(E_j\) according to the \(i\)-th row of \(H\).
% Each subproblem selects the top \(k\) experts by score. For example, $k=2$ routing sends each subproblem to the two highest‐scoring experts. 

Our CCO block employs a subproblem-level gating mechanism to dynamically route each context vector $h_c$ to a subset of experts from a pool consisting of 4 FF experts and 1 ID expert (thus $m=4$). Let $d$ be the hidden dimension and $W_G \in \mathbb{R}^{d \times (m+1)}$ denote the trainable gating weight matrix. Given a batch of context vectors $X = \{ h_c^b \}_{b=1}^B \in \mathbb{R}^{B \times d}$, where $B$ is the batch size, the gating network computes a score matrix: $H = X \cdot W_G \in \mathbb{R}^{B \times (m+1)}$, where each element $H_b^j$ represents the affinity or preference (score) of subproblem $b$ towards expert $j$.

For each subproblem, the router selects the Top-$k$ highest-scoring experts from the score vector $H_b^j$. These selected experts are then activated, and their outputs are weighted by a softmax-normalized version of their scores. In our POCCO with $k = 2$, each subproblem is routed to its two highest-scoring experts. 

This Top-$k$ routing plays a critical role by reducing computation, as only $k$ experts are activated per subproblem (i.e., sparse activation widely used in MoE structure~\cite{zhou2024mvmoe}). It also encourages expert specialization, since different subproblems (defined by context and weight vectors) tend to activate different subsets of experts. Additionally, it allows the inclusion of a parameter-free ID expert, which is often selected when minimal transformation is needed. This enables the router to learn when it is appropriate to leave a representation unchanged.

\section{Training Algorithm}
The training algorithm is provided in Algorithm~\ref{alg:po}. To train the model with preference learning, we first sample a batch of instances and weight vectors $\{(\mathcal{G}_b, \lambda_b)\}_{b=1}^{B}$ (as in Line 3). Then, for each scalarized subproblem $(\mathcal{G}_b, \lambda_b)$, we construct a set of winning–losing solution pairs $(\pi^{j,b}, \pi^{p,b}), \forall j,p\in \{1, \cdots, K\}$ (as in Lines 4-5). Training then proceeds by maximizing the likelihood of the winning solutions while minimizing that of the losing ones (as in Lines 6-7).

\label{appendix:alg}
\begin{algorithm}[!t]
    \caption{Preference-driven MOCO}
    \label{alg:po}
    \textbf{Input}: Instance distribution $\Tilde{\mathcal{G}}$, weight vector distribution $\Tilde{\lambda}$, number of training steps $E$, batch size $B$, number of tours $K$ per subproblem;\\
    % \textbf{Parameter}: Optional list of parameters\\
    \textbf{Output}: The trained policy network $\theta$;
    
    \begin{algorithmic}[1] %[1] enables line numbers
        % \STATE Let $t=0$.
        \STATE Initialize policy network $\theta$.
        \FOR{$e = 1$ to $E$}
        
        \STATE{$\lambda_{b} \sim $ \textsc{SampleWeightVector}($\Tilde{\lambda}$); $\mathcal{G}_b \sim $ \textsc{SampleInstance} ($\Tilde{\mathcal{G}}$),\quad $\forall b \in \left\{1, \cdots, B\right\}$}
        
        \STATE{$\pi^{j,b}\! \sim $\! \textsc{SampleSolutions}($p_{\theta}(\cdot|\mathcal{G}_b, \lambda_b)$), $\forall j \in \left\{1, \cdots, K\right\},\quad \forall b \in \left\{1, \cdots, B\right\}$}
        
        \STATE{$y_{j,p}^{b}\sim$ \textsc{PairwisePreference}$(1_{[\pi^{j,b}\lessdot \pi^{p,b}]}),\quad \forall j,p \in \{1,\cdots K\},\  \forall b \in \left\{1, \cdots, B\right\}$} 
        
        \STATE{Calculate gradient $\nabla_{\theta} \mathcal{L}(\theta)$ according to Eq.~(\ref{eq:po_loss})}
        
        \STATE{$\theta \leftarrow$ \text{ADAM}($\theta, \nabla_{\theta} \mathcal{L}(\theta)$)}
        \ENDFOR
    \end{algorithmic}
\end{algorithm}

\section{Decomposition Approaches}
The major decomposition techniques include weighted-sum, Tchebycheff, and penalty-based boundary intersection (PBI) approaches~\cite{zhang2007moea,miettinen2012nonlinear}, respectively.

\noindent\textbf{Weighted-sum Approach.} Given an MOCOP, the $i$th subproblem (i.e., SOCOP) is defined with the $i$th weight vector $\lambda_i$, such that,
\begin{equation}
    \min\quad g_{w}(\pi|\lambda_{i}) = \sum\nolimits_{j=1}^{\kappa} \lambda_{i}^{j} f_{j}(\pi),\,\, \pi\in \mathcal{X}
\end{equation}

% While the weighted-sum approach works well for convex Pareto fronts (PFs), it cannot be used for nonconvex PFs. 

\noindent\textbf{Tchebycheff Approach.} It minimizes the maximal distance between objectives and the ideal reference point, such that,
\begin{equation} \label{eq:Tchebycheff}
    \min \quad g_{t}(\pi|\lambda_{i}, z^{*}) = \max_{1\leq j \leq \kappa} \left\{\lambda_{i}^{j} | f_{j}(\pi) - z_{j}^{*}| \right\},\,\, \pi\in \mathcal{X}
\end{equation}
where $z^{*} =(z_1^*,\cdots,z_\kappa^*)^\top$ signifies the ideal reference point with $z_j^* \leq \min\left\{f_{j}(\pi)|\pi\in\mathcal{X}\right\}$.
% Unlike the weighted-sum approach, which is limited to convex Pareto fronts (PFs), the Tchebycheff approach is also effective for nonconvex PFs.
It guarantees that the optimal solution in Eq.~(\ref{eq:Tchebycheff}) under a specific (but unknown) weight vector $\lambda_{i}$ could be a Pareto optimal solution~\cite{choo1983proper}.

\noindent\textbf{PBI Approach.} This approach formulates the $i$th subproblem of an MOCOP as follows,
\begin{equation}
    \begin{array}{ll}
         \min & g_{p}(\pi|\lambda) = d_{1} + \alpha d_{2}   \\
          \rm{where} &  d_{1}\! =\! \frac{\left\|(F(\pi)\! - \!z^{*})^\top \lambda \right\|}{\left\| \lambda \right\|}  \\
                     &  d_{2}\! = \!\left\| F(\pi)\! -\! (z^{*}\! +\! d_{1} \lambda) \right\|, \pi\in \mathcal{X}
    \end{array}
\end{equation}
where $\alpha>0$ is a preset penalty item and $z^{*}$ is the ideal reference point as defined in the Tchebycheff approach. 
% The PBI approach can yield more uniformly distributed PFs than the Tchebycheff approach, especially in the case of more than two objectives~\cite{zhang2007moea}. 

\section{Hypervolume}
Hypervolume (HV) is a widely used indicator to evaluate approximate Pareto solutions to MOCOPs. Formally, the HV for a set of solutions $\mathcal{P}$ is defined as the volume of the subspace, which is weakly dominated by the solutions in $\mathcal{P}$ and bounded by a reference point $r^*$, such that,
\begin{equation}
    \text{HV}(\mathcal{P}) = \zeta^{\kappa}(\{r\in \mathbb{R}^{\kappa}|\exists\, \pi \in \mathcal{P}, \pi \prec r \prec r^*\}),
\end{equation}
where $\zeta^{\kappa}$ denotes the Lebesgue measure on the $\kappa$-dimensional space, i.e., the volume for a $\kappa$-dimensional subspace~\cite{while2006faster}. Since the range of objective values varies among different problems, we report the normalized HV $\Bar{H}(\mathcal{P})=\text{HV}(\mathcal{P})/\prod_{i=1}^{\kappa}|r^*_i-z_i|$, where the ideal point $z = (z_{1},\dots,z_{\kappa})$ satisfies $z_{i}
< 
\min\bigl\{f_{i}(\pi)\mid \pi \in\mathcal{P}\bigr\}
\bigl(\text{or }z_{i}>\max\{f_{i}(\pi)\mid \pi \in\mathcal{P}\}\text{ for maximization}\bigr),
\forall\,i\in\{1,\dots,\kappa\}$. All methods share the same $r^*$ and $z$ for an MOCOP, as given in Table~\ref{tab:ref_ideal}, and we report the average $\Bar{H}(\mathcal{P})$ over all test instances in this paper. 

\begin{table}[ht]
\centering
\caption{Reference points and ideal points.}
\label{tab:ref_ideal}
\begin{tabular}{@{} l c c c @{}}
\toprule
Problem   & Size & $r^*$               & $z$       \\ 
\midrule
\multirow{5}{*}{Bi-TSP}
          & 20  & (20,\,20)         & (0,\,0)   \\
          & 50  & (35,\,35)         & (0,\,0)   \\
          & 100 & (65,\,65)         & (0,\,0)   \\
          & 150 & (85,\,85)         & (0,\,0)   \\
          & 200 & (115,\,115)       & (0,\,0)   \\
\midrule
\multirow{3}{*}{Bi-CVRP}
          & 20  & (30,\,4)          & (0,\,0)   \\
          & 50  & (45,\,4)          & (0,\,0)   \\
          & 100 & (80,\,4)          & (0,\,0)   \\
\midrule
\multirow{3}{*}{Bi-KP}
          & 50  & (5,\,5)           & (30,\,30) \\
          & 100 & (20,\,20)         & (50,\,50) \\
          & 200 & (30,\,30)         & (75,\,75) \\
\midrule
\multirow{3}{*}{Tri-TSP}
          & 20  & (20,\,20,\,20)    & (0,\,0)   \\
          & 50  & (35,\,35,\,35)    & (0,\,0)   \\
          & 100 & (65,\,65,\,65)    & (0,\,0)   \\
\bottomrule
\end{tabular}
\end{table}

\section{Instance Augmentation} 
To further improve the performance of POCCO at the inference stage, we apply the instance augmentation proposed in~\cite{kwon2020pomo}, which is also used in PMOCO~\cite{lin2022pareto}. The rationale of instance augmentation is that an instance of Euclidean VRPs can be transformed into different ones that share the same optimal solution, e.g., by flipping coordinates for all nodes in an instance.  Given a coordinate $(x, y)$ in a VRP, there are eight simple transformations, i.e., $(x', y')$ = $(x, y); (y, x); (x, 1-y); (y, 1-x); (1-x, y); (1-y, x)$; $(1-x, 1-y); (1-y, 1-x)$. In our paper, we adopt these transformations for each objective, respectively. 
Hence, we could have 8 transformations for Bi-CVRP (since there is only one coordinate for each node), $8^2\!=\!64$ transformations for Bi-TSP, and $8^3\!=\!512$ transformations for Tri-TSP, respectively. 

\section{Impact of Neutralizing Non-Dominated Pairs}

Our method adopts a decomposition-based framework, where the MOCOP is divided into a set of scalarized subproblems, each associated with a weight vector. During training, the goal is to find high-quality solutions with respect to each subproblem, rather than globally exploring the entire Pareto front at once. In this context, preference learning operates at the subproblem level. We acknowledge, however, that sampled solutions may occasionally be mutually non-dominated in the original objective space. To investigate the effect of this scenario, we conducted an additional experiment where preference signals for such non-dominated pairs were set to 0, effectively treating them as equivalent. The results, shown in Table~\ref{tab:non-dominated}, indicate that this modification (NDS) significantly degrades model performance across all Bi-TSP instances.

This observation suggests that introducing indifference signals for non-dominated pairs hinders the learning process. One reason is that it reduces the amount of effective preference supervision per subproblem, especially in early training stages when many solutions are of similar quality. In contrast, using scalarized values provides a consistent and differentiable training signal aligned with the decomposition objective. Therefore, although scalarized comparisons may not reflect global Pareto dominance in every case, they remain well-justified and necessary within the decomposition-based learning paradigm, and do not appear to bias the overall Pareto front approximation based on our empirical findings.

\begin{table}[htbp]
    \centering
    \caption{Performance on Bi-TSP instances when neutralizing non-dominated pairs.}
    \label{tab:non-dominated}
    \begin{tabular}{c c c}
        \toprule
        \textbf{Problem Size} & \textbf{POCCO-W HV} & \textbf{NDS HV} \\
        \midrule
        Bi-TSP20  & 0.6275 & 0.5516 \\
        Bi-TSP50  & 0.6411 & 0.5160 \\
        Bi-TSP100 & 0.7055 & 0.5601 \\
        Bi-TSP150 & 0.7033 & 0.5494 \\
        Bi-TSP200 & 0.7371 & 0.5809 \\
        \bottomrule
    \end{tabular}
\end{table}

\section{Detailed Results on Benchmark Instances} \label{appendix: benchmark}
The detailed out-of-distribution generalization results are presented in Table~\ref{tab:kroab-results}, further confirming the exceptional generalization ability of our POCCO.

\begin{table}[t]
  \centering
  \caption{Performance on KroAB Instances}
  \label{tab:kroab-results}
  \begin{small}
    \renewcommand\arraystretch{0.7}
    \resizebox{0.98\textwidth}{!}{%
    \begin{tabular}{lccc ccc ccc}
      \toprule
      & \multicolumn{3}{c}{KroAB100}
      & \multicolumn{3}{c}{KroAB150}
      & \multicolumn{3}{c}{KroAB200} \\
      \cmidrule(lr){2-4}\cmidrule(lr){5-7}\cmidrule(lr){8-10}
      Method     & HV      & Gap       & Time   & HV      & Gap       & Time   & HV      & Gap       & Time    \\
      \midrule
       WS-LKH    & \textbf{0.7022} & -0.23\% & 2.3m   & \textbf{0.7017} & -0.59\% & 4.0m   & \textbf{0.7430} & -0.83\% & 5.6m   \\
       \midrule
    MOEA/D    & 0.6836 &  2.43\% & 5.8m   & 0.6710 &  3.81\% & 7.1m   & 0.7106 &  3.57\% & 7.3m   \\
    NSGA-II   & 0.6676 &  4.71\% & 7.0m   & 0.6552 &  6.08\% & 7.9m   & 0.7011 &  4.86\% & 8.4m   \\
    MOGLS     & 0.6817 &  2.70\% & 52m    & 0.6671 &  4.37\% & 1.3h   & 0.7083 &  3.88\% & 1.6h   \\
    PPLS/D-C  & 0.6785 &  3.15\% & 38m    & 0.6659 &  4.54\% & 1.4h   & 0.7100 &  3.65\% & 3.8h   \\
    \midrule
    DRL-MOA   & 0.6903 &  1.47\% & 10s    & 0.6794 &  2.61\% & 12s    & 0.7185 &  2.50\% & 18s    \\
    MDRL      & 0.6881 &  1.78\% & 9s     & 0.6831 &  2.08\% & 11s    & 0.7209 &  2.17\% & 16s    \\
    EMNH      & 0.6900 &  1.51\% & 9s     & 0.6832 &  2.06\% & 11s    & 0.7217 &  2.06\% & 16s    \\
    PMOCO     & 0.6878 &  1.83\% & 9s     & 0.6819 &  2.25\% & 12s    & 0.7193 &  2.39\% & 17s    \\
    \midrule
    CNH       & 0.6947 &  0.84\% & 16s    & 0.6892 &  1.20\% & 19s    & 0.7250 &  1.61\% & 22s    \\
    \textbf{POCCO-C}   & 0.6965 &  0.59\% & 30s    & 0.6925 &  0.73\% & 40s    & 0.7302 &  0.91\% & 50s    \\
    \midrule
    WE-CA     & 0.6948 &  0.83\% & 9s     & 0.6924 &  0.75\% & 12s    & 0.7317 &  0.71\% & 16s    \\
    \textbf{POCCO-W}   & 0.6981 &  0.36\% & 20s    & 0.6946 &  0.43\% & 31s    & 0.7345 &  0.33\% & 40s    \\
    \midrule
    MDRL-Aug  & 0.6950 &  0.80\% & 10s    & 0.6890 &  1.23\% & 16s    & 0.7261 &  1.47\% & 25s    \\
    EMNH-Aug  & 0.6958 &  0.69\% & 10s    & 0.6892 &  1.20\% & 16s    & 0.7270 &  1.34\% & 25s    \\
    PMOCO-Aug & 0.6937 &  0.98\% & 11s    & 0.6886 &  1.29\% & 18s    & 0.7251 &  1.60\% & 30s    \\
    \midrule
    CNH-Aug   & 0.6980 &  0.37\% & 17s    & 0.6938 &  0.54\% & 26s    & 0.7303 &  0.90\% & 37s    \\
    \textbf{POCCO-C-Aug} & 0.6999 & 0.10\% & 32s  & 0.6959 &  0.24\% & 48s    & 0.7334 &  0.47\% & 1.1m   \\
    \midrule
    WE-CA-Aug & 0.6990 &  0.23\% & 10s    & 0.6957 &  0.27\% & 20s    & 0.7349 &  0.27\% & 31s    \\
    \textbf{POCCO-W-Aug} & \underline{0.7006} & 0.00\% & 22s  & \underline{0.6976} &  0.00\% & 39s    & \underline{0.7369} &  0.00\% & 59s    \\
      \bottomrule
    \end{tabular}}%
  \end{small}
\end{table}

\begin{table}[t]
  \centering
  \caption{Performance on Bi‐TSP150 and Bi‐TSP200 Instances}
  \label{tab:bitsp150-200}
  \begin{small}
  \renewcommand\arraystretch{0.7}
  \begin{tabular}{l ccc ccc}
    \toprule
    & \multicolumn{3}{c}{Bi‐TSP150} & \multicolumn{3}{c}{Bi‐TSP200} \\
    \cmidrule(lr){2-4} \cmidrule(lr){5-7}
    Method & HV & Gap & Time & HV & Gap & Time \\
    \midrule
   WS-LKH        & \textbf{0.7149} & –1.23\% & 13h   & \textbf{0.7490} & –1.23\% & 22h   \\
   \midrule
    MOEA/D        & 0.6809 &  3.58\% & 2.4h  & 0.7139 &  3.51\% & 2.7h  \\
    NSGA-II       & 0.6659 &  5.71\% & 6.8h  & 0.7045 &  4.78\% & 6.9h  \\
    MOGLS         & 0.6768 &  4.16\% & 22h   & 0.7114 &  3.85\% & 38h   \\
    PPLS/D-C      & 0.6784 &  3.94\% & 21h   & 0.7106 &  3.96\% & 32h   \\
    \midrule
    DRL-MOA       & 0.6901 &  2.28\% & 36s   & 0.7219 &  2.43\% & 1.2m  \\
    MDRL          & 0.6922 &  1.98\% & 36s   & 0.7251 &  2.00\% & 1.1m  \\
    EMNH          & 0.6930 &  1.87\% & 37s   & 0.7260 &  1.88\% & 1.1m  \\
    PMOCO         & 0.6910 &  2.15\% & 42s   & 0.7231 &  2.27\% & 1.3m  \\
    \midrule
    CNH           & 0.6985 &  1.09\% & 50s   & 0.7292 &  1.45\% & 1.4m  \\
    \textbf{POCCO-C}       & 0.7011 &  0.72\% & 1.5m  & 0.7333 &  0.89\% & 2.5m  \\
    \midrule
    WE-CA         & 0.7008 &  0.76\% & 45s   & 0.7346 &  0.72\% & 1.3m  \\
    \textbf{POCCO-W}       & 0.7033 &  0.41\% & 1.4m  & 0.7371 &  0.38\% & 2.4m  \\
    \midrule
    MDRL-Aug      & 0.6976 &  1.22\% & 37m   & 0.7299 &  1.35\% & 1.1h  \\
    EMNH-Aug      & 0.6983 &  1.12\% & 39m   & 0.7307 &  1.24\% & 1.1h  \\
    PMOCO-Aug     & 0.6967 &  1.35\% & 40m   & 0.7283 &  1.57\% & 1.2h  \\
    \midrule
    CNH-Aug       & 0.7025 &  0.52\% & 41m   & 0.7343 &  0.76\% & 1.2h  \\
    \textbf{POCCO-C-Aug}   & 0.7043 &  0.27\% & 55m   & 0.7366 &  0.45\% & 1.5h  \\
    \midrule
    WE-CA-Aug     & 0.7044 &  0.25\% & 42m   & 0.7381 &  0.24\% & 1.2h  \\
    \textbf{POCCO-W-Aug}   & \underline{0.7062} &  0.00\% & 45m   & \underline{0.7399} &  0.00\% & 1.4h  \\
    \bottomrule
  \end{tabular}
  \end{small}
\end{table}

\begin{table}[htbp]
    \centering
    \caption{HV on large Bi-TSP instances.}
    \label{tab:large_tsp}
    \begin{tabular}{c c c}
        \toprule
        \textbf{Problem Size} & \textbf{WE-CA} & \textbf{POCCO-W} \\
        \midrule
        Bi-TSP300  & 0.7441 & \textbf{0.7458} \\
        Bi-TSP500  & 0.7476 & \textbf{0.7592} \\
        Bi-TSP1000 & 0.7186 & \textbf{0.7400} \\
        \bottomrule
    \end{tabular}
\end{table}

\begin{table}[htbp]
    \centering
    \caption{HV on large MOKP instances.}
    \label{tab:large_kp}
    \begin{tabular}{c c c}
        \toprule
        \textbf{Problem Size} & \textbf{WE-CA} & \textbf{POCCO-W} \\
        \midrule
        Bi-KP300  & \textbf{0.6000} & 0.5947 \\
        Bi-KP500  & 0.4244 & \textbf{0.5658} \\
        Bi-KP1000 & 0.2439 & \textbf{0.8408} \\
        \bottomrule
    \end{tabular}
\end{table}
% \begin{table}[htbp]
%     \centering
%     \begin{minipage}{0.48\textwidth}
%         \centering
%         \caption{HV on large Bi-TSP instances.}
%         \label{tab:large_tsp}
%         \begin{tabular}{c c c}
%             \toprule
%             \textbf{Problem Size} & \textbf{WE-CA} & \textbf{POCCO-W} \\
%             \midrule
%             Bi-TSP300  & 0.7441 & \textbf{0.7458} \\
%             Bi-TSP500  & 0.7476 & \textbf{0.7592} \\
%             Bi-TSP1000 & 0.7186 & \textbf{0.7400} \\
%             \bottomrule
%         \end{tabular}
%     \end{minipage}
%     \hfill
%     \begin{minipage}{0.48\textwidth}
%         \centering
%         \caption{HV on large MOKP instances.}
%         \label{tab:large_kp}
%         \begin{tabular}{c c c}
%             \toprule
%             \textbf{Problem Size} & \textbf{WE-CA} & \textbf{POCCO-W} \\
%             \midrule
%             Bi-KP300  & \textbf{0.6000} & 0.5947 \\
%             Bi-KP500  & 0.4244 & \textbf{0.5658} \\
%             Bi-KP1000 & 0.2439 & \textbf{0.8408} \\
%             \bottomrule
%         \end{tabular}
%     \vspace{-3mm}
%     \end{minipage}
% \end{table}

\section{Experimental Results on the Larger Problem Sizes}
The results on Bi-TSP150/200, summarized in Table~\ref{tab:bitsp150-200}, show that POCCO-W consistently achieves the best generalization performance compared to all neural baselines and classical MOEAs.
\label{appendix:large_size}

Furthermore, to demonstrate scalability, we include experimental results on even larger problems, including Bi-TSP with 300/500/1000 nodes and MOKP with 300/500/1000 items. These results are summarized in Table~\ref{tab:large_tsp} and Table~\ref{tab:large_kp}. As shown, POCCO-W consistently outperforms the baseline WE-CA in most large-scale cases.

\section{Comparison with DPO}

We have added a comparative experiment between our preference learning (PL) method and the recent DPO objective, as shown in Table~\ref{tab:dpo}. Since DPO requires two models, i.e., a policy model and a reference model, we use the same architecture and initialization for both, with the reference model updated from the previous epoch. This setup significantly increases training time for DPO (122h vs. 36h for POCCO-W).

As shown in Table~\ref{tab:dpo}, while DPO achieves slightly better performance on the smallest instance (MOTSP20), it is consistently outperformed by POCCO-W on larger instances (MOTSP50–200). These results suggest that our PL method offers a more favorable trade-off in terms of both training efficiency and solution quality, especially for large-scale problems.

\begin{table}[htbp]
    \centering
    \caption{HV of different PL methods on Bi-TSP instances.}
    \label{tab:dpo}
    \begin{tabular}{c c c}
        \toprule
        \textbf{Problem Size} & \textbf{POCCO-W} & \textbf{DPO} \\
        \midrule
        Bi-TSP20  & 0.6275 & \textbf{0.6276} \\
        Bi-TSP50  & \textbf{0.6411} & 0.6400 \\
        Bi-TSP100 & \textbf{0.7055} & 0.7030 \\
        Bi-TSP150 & \textbf{0.7033} & 0.6998 \\
        Bi-TSP200 & \textbf{0.7371} & 0.7331 \\
        \bottomrule
    \end{tabular}
\end{table}

\section{Hyperparameter Study}
\label{appendix: hyperparam}

\begin{figure}[htbp]
  \centering
  % 第一行，两张图
  \begin{subfigure}[b]{0.49\textwidth}
    \includegraphics[width=\textwidth]{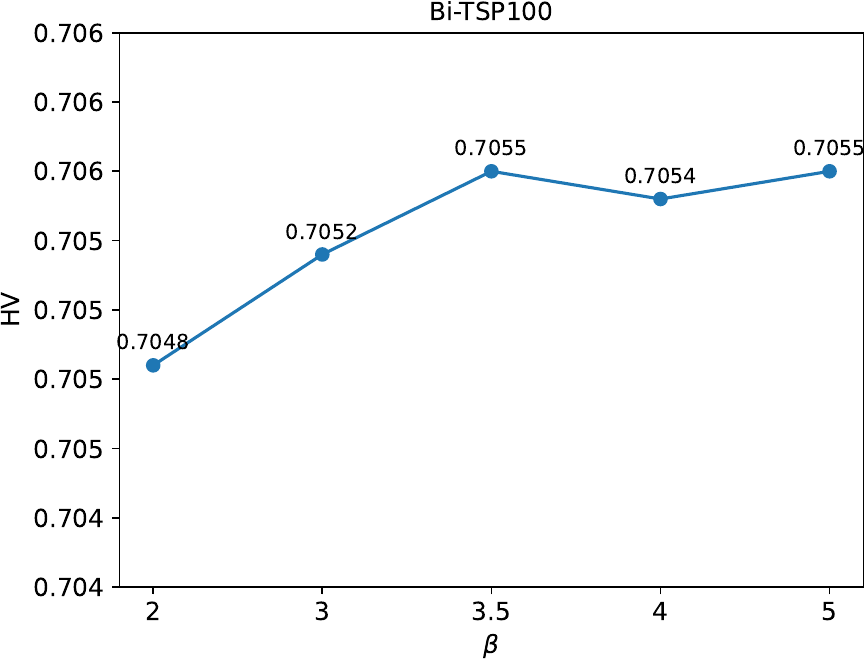}
  \end{subfigure}
  \hfill
  \begin{subfigure}[b]{0.49\textwidth}
    \includegraphics[width=\textwidth]{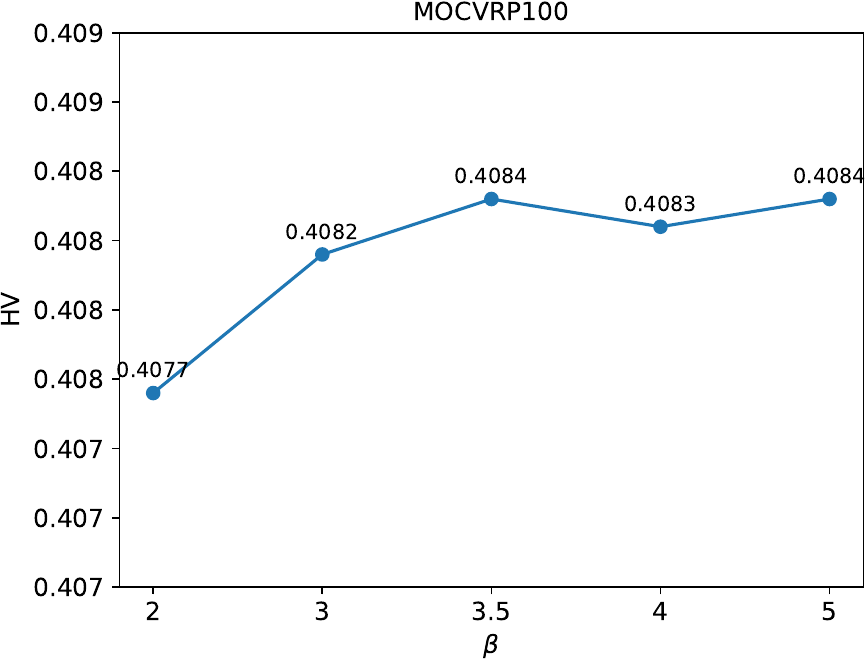}
  \end{subfigure}

  \vspace{1em} % 调整上下行的间距

  % 第二行，两张图
  \begin{subfigure}[b]{0.49\textwidth}
    \includegraphics[width=\textwidth]{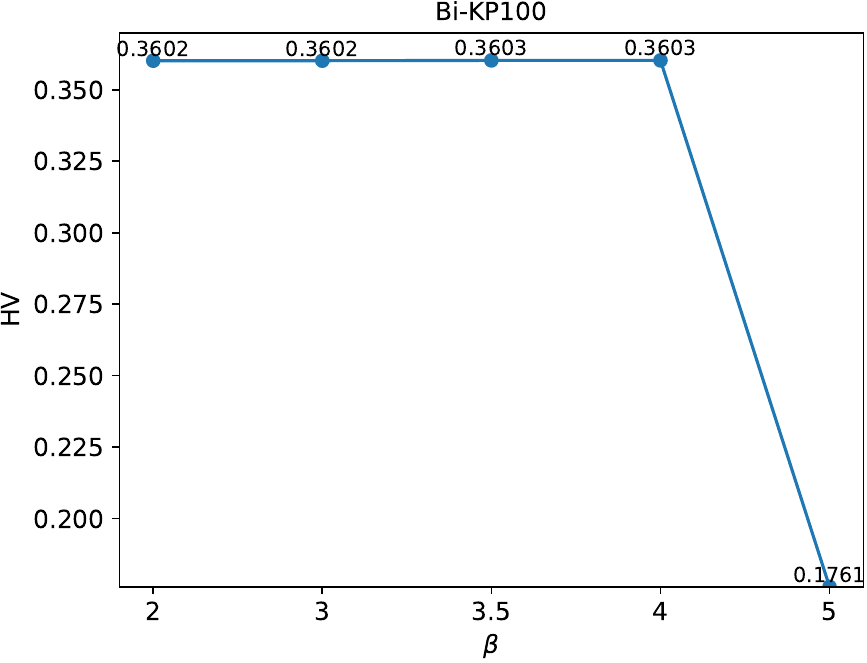}
  \end{subfigure}
  \hfill
  \begin{subfigure}[b]{0.49\textwidth}
    \includegraphics[width=\textwidth]{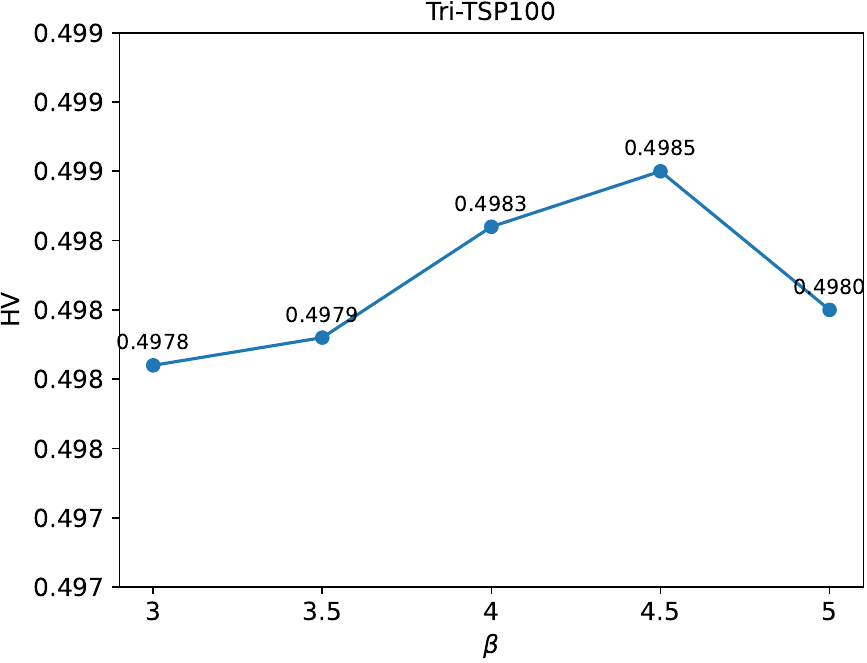}
  \end{subfigure}

  \caption{Effectis of the $\beta$.}
  \label{fig:four_images}
\end{figure}

\noindent\textbf{Effects of the $\beta$.}
Fig.\ref{fig:four_images} shows how the temperature parameter $\beta$ in the preference learning loss affects performance (HV) across four benchmark tasks. A moderate value of $\beta$ consistently yields the best results. For the three bi-objective problems (Bi-TSP100, MOCVRP100, Bi-KP100), performance peaks around $\beta=3.5$; larger or smaller values provide no additional benefit, and in Bi-KP100, an overly large $\beta=5$ sharply degrades HV. For the tri-objective problem (Tri-TSP100), performance improves up to $\beta=4.5$ and then declines slightly. Overall, these trends justify the default settings adopted in our experiments: $\beta=3.5$ for bi-objective tasks and $\beta=4.5$ for tri-objective tasks.

\noindent\textbf{Effects of the number of CCO layers.}
The CCO block sits in the decoder, whose sequence length T grows with problem size. Each additional CCO layer, therefore, adds a full set of gating and expert computations at every decoding step. Hence, stacking too many CCO layers can inflate the runtime disproportionately.
Table \ref{tab:cco_layers} reports the trade-off. Using two CCO layers yields the highest HV on both Bi-TSP100 and Bi-TSP200, but increases inference time by roughly 60\% and 130 \%, respectively. A third layer further slows inference while slightly reducing HV. We therefore adopt a single-layer CCO as the default, which preserves most of the performance gain while keeping computation modest. For larger instances or latency-sensitive applications, the one-layer setting offers a favorable balance between solution quality and speed.

 \begin{table}[hbt]
  \centering
  \caption{Effects of the number of CCO block layers.}
  \begin{tabular}{l cc cc}
    \toprule
    & \multicolumn{2}{c}{Bi-TSP100} & \multicolumn{2}{c}{Bi-TSP200} \\
    \cmidrule(lr){2-3} \cmidrule(lr){4-5} 
    Method     & HV     & Time         & HV     & Time         \\
    \midrule
    POCCO-W       & 0.7055 & 36\,s   & 0.7371 & 2.4\,m  \\
    2 CCO layers & 0.7058 & 59\,s   & 0.7379 & 5.6\,m  \\
    3 CCO layers & 0.7049 & 85\,s   & 0.7352 & 7.9\,m  \\
    \bottomrule
  \end{tabular}
  \label{tab:cco_layers}
\end{table}

\noindent \textbf{Effects of the number of experts.}
We have added an ablation study to analyze the impact of the number of experts in Table~\ref{tab:expert_num}. Specifically, POCCO-W (5E) refers to the original model presented in the paper, which includes 4 feedforward (FF) experts and 1 parameter-free identity (ID) expert. We additionally evaluate four variants:
\begin{enumerate}
    \item POCCO-W (3E): 2 FF + 1 ID;
    \item POCCO-W (9E): 8 FF + 1 ID;
    \item POCCO-W (9E\_2D): 8 FF + 1 ID, trained on twice the amount of data;
    \item POCCO-W (17E): 16 FF + 1 ID.
\end{enumerate}

All variants are trained under the same settings as POCCO-W (5E) for a fair comparison, except for POCCO-W (9E\_2D), which is trained on more data. 

As shown in Table~\ref{tab:expert_num}, all POCCO-W variants outperform the backbone model WE-CA, confirming the effectiveness of the expert-based architecture. Moreover, POCCO-W (3E) and POCCO-W (5E) achieve better overall performance than POCCO-W (9E) and POCCO-W (17E), the latter of which requires additional data scaling to realize performance gains. To strike a better balance between computational cost and solution quality, we select POCCO-W (5E) as our default model.

\begin{table}[htbp]
    \centering
    \scriptsize
    \caption{Effects of the number of experts.}
    \label{tab:expert_num}
    \begin{tabular}{c c c c c c c}
        \toprule
        \textbf{Problem Size} & \textbf{WE-CA} & \textbf{POCCO-W (3E)} & \textbf{POCCO-W (5E)} & \textbf{Exp\_num (9E)} & \textbf{Exp\_num (9E\_2D)} & \textbf{Exp\_num (17E)} \\
        \midrule
        Bi-TSP20  & 0.6270 & \textbf{0.6275} & \textbf{0.6275} & \textbf{0.6275} & \textbf{0.6275} & \textbf{0.6275} \\
        Bi-TSP50  & 0.6392 & 0.6410 & \textbf{0.6411} & 0.6408 & \textbf{0.6411} & 0.6408 \\
        Bi-TSP100 & 0.7034 & \textbf{0.7057} & 0.7055 & 0.7050 & 0.7053 & 0.7050 \\
        Bi-TSP150 & 0.7008 & \textbf{0.7035} & 0.7033 & 0.7026 & 0.7031 & 0.7027 \\
        Bo-TSP200 & 0.7346 & 0.7363 & \textbf{0.7371} & 0.7364 & 0.7370 & 0.7361 \\
        \bottomrule
    \end{tabular}
\end{table}

\begin{figure}[htbp]
    \centering
    \includegraphics[width=0.75\textwidth]{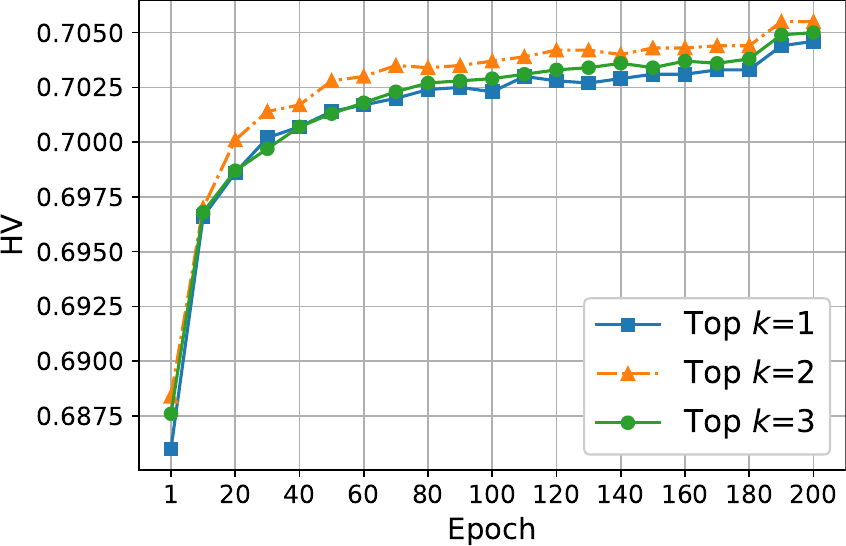}
    % \vspace{-3mm}
    \caption{Effects of $\operatorname{Top}k$.}
    \label{fig:effect_k}
\end{figure}

\noindent\textbf{Effects of $\operatorname{Top}k$.}
Fig.\ref{fig:effect_k} plots validation HV on Bi-TSP100 for $k\!\in\!\{1,2,3\}$. Selecting two experts per token ($k$=2) converges fastest and attains the highest final HV. A single expert ($k$=1) limits model capacity, leading to slower early progress and a slightly lower plateau. Choosing three experts ($k$=3) increases computation relative to $k$=2 yet offers no benefit and even marginally reduces HV, likely because the additional expert dilutes specialization and weakens sparsity. We therefore adopt $k$=2 as the default, balancing performance and efficiency.

\section{Experimental Results of Scalarized Subproblems}
To assess subproblem optimality, we report scalarized objective values for three representative weight vectors $\lambda=(1,0),\,(0.5,0.5),\,(0,1)$ on Bi-TSP100 (Table \ref{tab:subproblem_results}). We also compare against single-objective solvers LKH and POMO, each tuned to the corresponding subproblem. Among neural MOCOP methods, POCCO-W attains the smallest optimality gaps across all weight settings. 
% In contrast, WE-CA-Aug shows larger gaps—most pronounced at $\lambda=(0.5,0.5)$.
Ablating CCO (POCCO-Aug w/o CCO) increases optimality gaps across all settings, and eliminating preference learning as well (WE-CA-Aug) further degrades performance. Notably, POCCO-W-Aug even outperforms POMO-Aug on the subproblem with $\lambda=(0,1)$.
% Overall, full POCCO-W combines these strengths and attains the best aggregate performance on the scalarized subproblems.

\begin{table}[ht]
\centering
\caption{Performance comparison under different weight settings $(\omega_1,\omega_2)$.}
\begin{tabular}{l  cc  cc  cc}
\toprule
 & \multicolumn{2}{c}{$\lambda=(1,0)$} & \multicolumn{2}{c}{$\lambda=(0.5,0.5)$} & \multicolumn{2}{c}{$\lambda=(0,1)$} \\
\cmidrule(lr){2-3} \cmidrule(lr){4-5} \cmidrule(lr){6-7}
Method & Obj & Gap & Obj & Gap & Obj & Gap \\
\midrule
WS-LKH              &  7.7632 & 0.00\% & 17.3094 & 0.00\% &  7.7413 & 0.00\% \\
POMO-Aug            &  7.7659 & 0.03\% & 17.4421 & 0.77\% &  7.7716 & 0.39\% \\
WE-CA-Aug           &  7.8132 & 0.64\% & 17.4994 & 1.10\% &  7.7888 & 0.61\% \\
POCCO-Aug w/o CCO    &  7.7970 & 0.44\% & 17.4629 & 0.89\% &  7.7772 & 0.46\% \\
POCCO-W-Aug           &  7.7827 & 0.25\% & 17.4460 & 0.79\% &  7.7602 & 0.24\% \\
\bottomrule
\end{tabular}
\label{tab:subproblem_results}
\end{table}

\section{Comparison with Tchebycheff Scalarization}

Intuitively, our POCCO framework is applicable to any MOCOP where the objectives can be scalarized using decomposition-based techniques, including weighted-sum (WS), Tchebycheff (TCH), and penalty-based boundary intersection (PBI). This is because the preference signal between two solutions is determined solely by the ordering of their scalarized objective values, regardless of the specific scalarization method used.

To empirically demonstrate this generality, we added an experiment comparing POCCO-W trained with WS and TCH scalarization methods. As shown in Table~\ref{tab:TCH}, WS consistently outperforms TCH on Bi-TSP50 to Bi-TSP200, indicating that WS offers more robust performance across different problem sizes. This observation is also consistent with findings in~\cite{chen2025rethinking}, which report that WS—despite its simplicity—often outperforms TCH in the studied MOCOP setting. In general, however, WE-CA (TCH) performs worse than WE-CA (WS), which achieves a score of 0.6392, and is further outperformed by POCCO-W (TCH), which achieves 0.6395 on Bi-TSP50.

\begin{table}[htbp]
    \centering
    \caption{HV of Different Decomposition Approaches on Bi-TSP Instances.}
    \label{tab:TCH}
    \begin{tabular}{c c c c}
        \toprule
        \textbf{Problem Size} & \textbf{WE-CA (WS)} & \textbf{POCCO-W (WS)} & \textbf{POCCO-W (TCH)} \\
        \midrule
        Bi-TSP50   & 0.6392 & 0.6411 & 0.6395 \\
        Bi-TSP100  & 0.7034 & 0.7055 & 0.7031 \\
        Bi-TSP150  & 0.7008 & 0.7033 & 0.6998 \\
        Bi-TSP200  & 0.7346 & 0.7371 & 0.7329 \\
        \bottomrule
    \end{tabular}
\end{table}

\section{GPU Memory Usage}

\begin{table}[htbp]
    \centering
    \caption{Comparison of GPU Memory Usage.}
    \label{tab:gpu}
    \begin{tabular}{c c c c}
        \toprule
        \textbf{Problem Size} & \textbf{WE-CA} & \textbf{CNH} & \textbf{POCCO-W} \\
        \midrule
        Bi-TSP50   & 1051 MB & 1077 MB & 1849 MB \\
        MOCVRP50   & 1790 MB & 1501 MB & 4522 MB \\
        Bi-KP50    & 916 MB  & 916 MB  & 2417 MB \\
        Tri-TSP50  & 1052 MB & 1077 MB & 2383 MB \\
        \bottomrule
    \end{tabular}
\end{table}

 We present a comparison of GPU memory usage between POCCO-W (ours), the backbone model (WE-CA), and CNH in Table~\ref{tab:gpu}. All models are trained across problem instances with size $n =50$. As shown in Table~\ref{tab:gpu}, POCCO-W incurs higher GPU memory usage, approximately doubling that of the baselines WE-CA and CNH. However, this increased resource cost is accompanied by significant improvements in solution quality, as demonstrated in our experimental results.

\section{Summary of Decomposition-based Neural MOCOP Solvers}

Table \ref{tab:decomp_methods} compares key features of decomposition-based neural MOCOP solvers. POCCO, which establishes new SOTA performance, is a plug-and-play framework that augments any existing solver. It inserts a CCO block that learns a diverse ensemble of policies, adding parameters yet surpassing prior methods. Crucially, each subproblem activates only two experts (one of which may be a parameter-free identity expert), so the extra computational load remains minimal. POCCO also employs a pairwise preference learning approach that further boosts performance without introducing additional parameters.

\begin{table}[ht]
\centering
\caption{Summary of the decomposition-based neural MOCOP solvers.}
\label{tab:decomp_methods}
% \resizebox{0.89\textwidth}{!}{ 
\begin{tabular}{@{} l l l l @{}}
\toprule
Method   & Learning method  & Paradigm   & \#Parameters                           \\ 
\midrule
DRL-MOA  & Transfer learning+RL                 & one-to-one   & 133.37M                                 \\
MDRL     & Meta learning+RL                    & one-to-one   & 133.37M                                \\
EMNH     & Meta learning+RL                   & one-to-one   & 133.37M                                   \\
PMOCO    & RL                      & one-to-many  & 1.50M                                     \\
CNH      & RL     & one-to-many  & 1.63M        \\    
\textbf{POCCO-C} & Preference learning &
one-to-many & 2.16M \\
WE-CA    & RL    & one-to-many  & 1.47M        \\
\textbf{POCCO-W} & Preference learning & one-to-many &
2.00M \\
\bottomrule
\end{tabular}
% }
\end{table}

\section{Broader Impacts}
\label{appendix:impact}
%This paper focuses on real-world scenarios and proposes a novel Proactive Infeasibility Prevention (PIP) framework to enhance the capabilities of neural methods towards solving more complex VRPs. Potential positive societal impacts include: 1) enhancing industrial efficiency, e.g., in logistics and transportation. By preemptively identifying infeasible solutions, it can reduce computational overheads and improve the efficiency of decision-making process; 2) advancing the AI and operation research (OR) communities. Our PIP framework aims to alleviate the existing challenges in the neural VRP solvers, thereby promoting the advancement of AI as well as OR. On the other hand, negative societal impacts may include environmental unfriendliness due to computational resource usage.

POCCO offers several positive societal impacts. By dynamically routing computation and learning from preference signals, it accelerates multi-objective decision-making in logistics, manufacturing, and energy planning, reducing costly trial-and-error loops and boosting operational efficiency. Its conditional-computation design activates only the required network capacity for each subproblem, cutting FLOP counts and energy consumption relative to dense models of comparable accuracy. Finally, by encouraging exploration and adaptive capacity allocation, POCCO broadens the applicability of neural solvers in complex optimization tasks, advancing both AI and operations research and enabling practitioners to tackle larger, real-world problems with fewer computational resources.

\section{Licenses for Existing Assets}
\label{appendix:assets}
% The used assets in this work are listed in Table \ref{tab:asset}, which are all open-source for academic research. We have released our source code with the MIT License.

The used assets in this work are listed in Table 17. Where applicable, we reference publicly available implementations for evaluation or reproduction purposes. Our source code is released under the MIT License.

\begin{table}[htbp]
  \centering
  \caption{Used assets, licenses, and their usage.}
  \vspace{4pt}
  \begin{threeparttable}
    \resizebox{0.9\textwidth}{!}{
    \begin{tabular}{c|c|c|c}
    \toprule
    \toprule
    % \midrule
    \textbf{Type}  & \textbf{Asset} & \textbf{License} & \textbf{Usage} \\
    \midrule
    % \midrule
    \multirow{5}[2]{*}{Code} & LKH \cite{helsgaun2000effective} & Available for academic use & Evaluation \\
     & DRL-MOA \cite{likaiwen2020deep} & No license (assumed all rights reserved) & Evaluation \\
      & MDRL \cite{zhang2022meta} & No license (assumed all rights reserved) & Evaluation \\
      & EMNH \cite{chen2024efficient} & No license (assumed all rights reserved)
 & Evaluation \\
     & PMOCO \cite{lin2022pareto} & MIT License
 & Evaluation \\
      & CNH \cite{fan2024conditional} & MIT License & Revision \\
      & WE-CA \cite{chen2025rethinking}   &  MIT License& Revision \\
      % & MVMoE \cite{zhou2024mvmoe}   &  MIT License& Revision \\
    \midrule
    % \multirow{1}[2]{*}{Datasets}
     Datasets & Chen et al.\cite{chen2025rethinking} &  MIT License & Evaluation  \\
    %   & \citet{todosijevic2017general} & Available for academic use \\
    %   & \citet{da2010general} & Available for academic use \\
    %   & \citet{da2010general} & Available for academic use \\
    % \midrule
    \bottomrule
    \bottomrule
    \end{tabular}}
    \end{threeparttable}
  \label{tab:asset}%
\end{table}%

% \newpage
% \bibliographystyle{plain}
% \bibliography{ref}

% \end{document}

\end{document}